\documentclass[12pt]{article}
\usepackage{arxiv}

\usepackage[utf8]{inputenc} 
\usepackage[T1]{fontenc}    
\usepackage{hyperref}       
\hypersetup{
    colorlinks=true,   
    citecolor=black,   
    linkcolor=black,   
    urlcolor=blue      
}

\usepackage{amsmath}
\usepackage{url}            
\usepackage{booktabs}       
\usepackage{amsfonts}       
\usepackage{nicefrac}       
\usepackage{microtype}      
\usepackage{lipsum}
\usepackage{graphicx}
\graphicspath{ {./images/} }
\usepackage{longtable}
\usepackage{booktabs}

\newcommand\blfootnote[1]{%
  \begingroup
  \renewcommand\thefootnote{}\footnote{#1}%
  \addtocounter{footnote}{-1}%
  \endgroup
}

\usepackage{fancyhdr}
\usepackage[numbers,sort&compress]{natbib}
\bibpunct[, ]{[}{]}{,}{n}{,}{,}
\renewcommand\bibfont{\fontsize{10}{12}\selectfont}
\usepackage{rotating}
\usepackage{array}

\author{
 Sudip Chakrabarty \\
 School of Computer Engineering, KIIT University\\
  \texttt{sudipchakrabarty6@gmail.com} \\
  }

\begin{document}

\title{YOLO26: An Analysis of NMS-Free End to End Framework for Real-Time Object Detection}

\maketitle
\blfootnote{This article presents a secondary analytical review of YOLO26 based exclusively on publicly available documentation, benchmarks, and technical descriptions released by Ultralytics. For official documentation of YOLO26, visit: \url{https://docs.ultralytics.com/models/yolo26/}}

\begin{abstract}
The ``You Only Look Once'' (YOLO) framework has long served as a standard for real-time object detection, though traditional iterations have utilized Non-Maximum Suppression (NMS) post-processing, which introduces specific latency and hyperparameter variables. This paper presents a comprehensive architectural analysis of YOLO26, a model that shifts toward a native end-to-end learning strategy by eliminating NMS. This study examines the core mechanisms driving this framework: the MuSGD optimizer for backbone stabilization, Small-Target-Aware Label Assignment (STAL), and ProgLoss for dynamic supervision. To contextualize its performance, this article reviews exhaustive benchmark data from the COCO \texttt{val2017} leaderboard. This evaluation provides an objective comparison of YOLO26 across various model scales (Nano to Extra-Large) against both prior CNN lineages and contemporary Transformer-based architectures (e.g., RT-DETR, DEIM, RF-DETR), detailing the observed speed-accuracy trade-offs and parameter requirements without asserting a singular optimal model. Additionally, the analysis covers the framework's unified multi-task capabilities, including the YOLOE-26 open-vocabulary module for promptable detection. Ultimately, this paper serves to document how decoupling representation learning from heuristic post-processing impacts the "Export Gap" and deterministic latency in modern edge-based computer vision deployments.

\vspace{0.5em}
\noindent \textbf{Keywords:} YOLO26, End-to-End Object Detection, NMS-Free, MuSGD, ProgLoss, YOLOE-26, Open-Vocabulary Detection, Real-Time Computer Vision.
\end{abstract}


\section{Introduction}
\label{sec:intro}
Computer vision has evolved rapidly from basic image processing techniques such as edge detection and morphological filtering into a domain dominated by deep learning. At the forefront of this evolution is Object Detection, the fundamental task of identifying and localizing instances of semantic objects within a digital image \cite{zou2023object,zhao2019object}. Unlike simple classification, which assigns a single label to an image, object detection requires the simultaneous prediction of class labels and geometric bounding boxes. This capability is the cornerstone of modern automation, underpinning critical applications ranging from autonomous driving and robotic navigation to medical image analysis and real-time surveillance \cite{jiao2019survey}. As the demand for real-time analysis has grown, the field has shifted away from computationally heavy two-stage detectors (like Faster R-CNN) toward efficient one-stage architectures that prioritize inference speed without compromising accuracy \cite{ren2015faster,liu2016ssd}.

\subsection{The Ultralytics Legacy}
In this landscape, Ultralytics has emerged as the defining force in real-time detection. Beginning with the standardization of the YOLO (You Only Look Once) architecture, Ultralytics has consistently pushed the boundaries of efficiency. Their iterative releases—most notably YOLOv5 \cite{jocher2020yolov5} and YOLOv8 \cite{sohan2024review} —stablished  a new industry standard by combining Cross-Stage Partial (CSP) backbones with user-friendly deployment pipelines. These models successfully democratized AI, allowing complex detection tasks to run on edge devices with limited computational resources. However, even these state-of-the-art models largely relied on Non-Maximum Suppression (NMS) post-processing, a sequential step that introduces latency variability in dense scenes.

\subsection{YOLO26: Redefining Real-Time Edge Inference}
Released in \textbf{January 2026}, YOLO26 establishes a new milestone in the history of real-time object detection. To quantify this leap, the \textbf{Ultralytics team} has released official benchmarks comparing YOLO26 against a comprehensive suite of predecessors (YOLOv5 \cite{jocher2020yolov5} through YOLO11 \cite{jocher2024yolo11,chakrabartyadvance}) and competitive architectures such as RTMDet \cite{lyu2022rtmdet}, DAMO-YOLO \cite{xu2022damoyolo}, and PP-YOLOE+ \cite{xu2022ppyoloe}.

\begin{figure}[h!]
    \centering
    \includegraphics[width=1.0\linewidth]{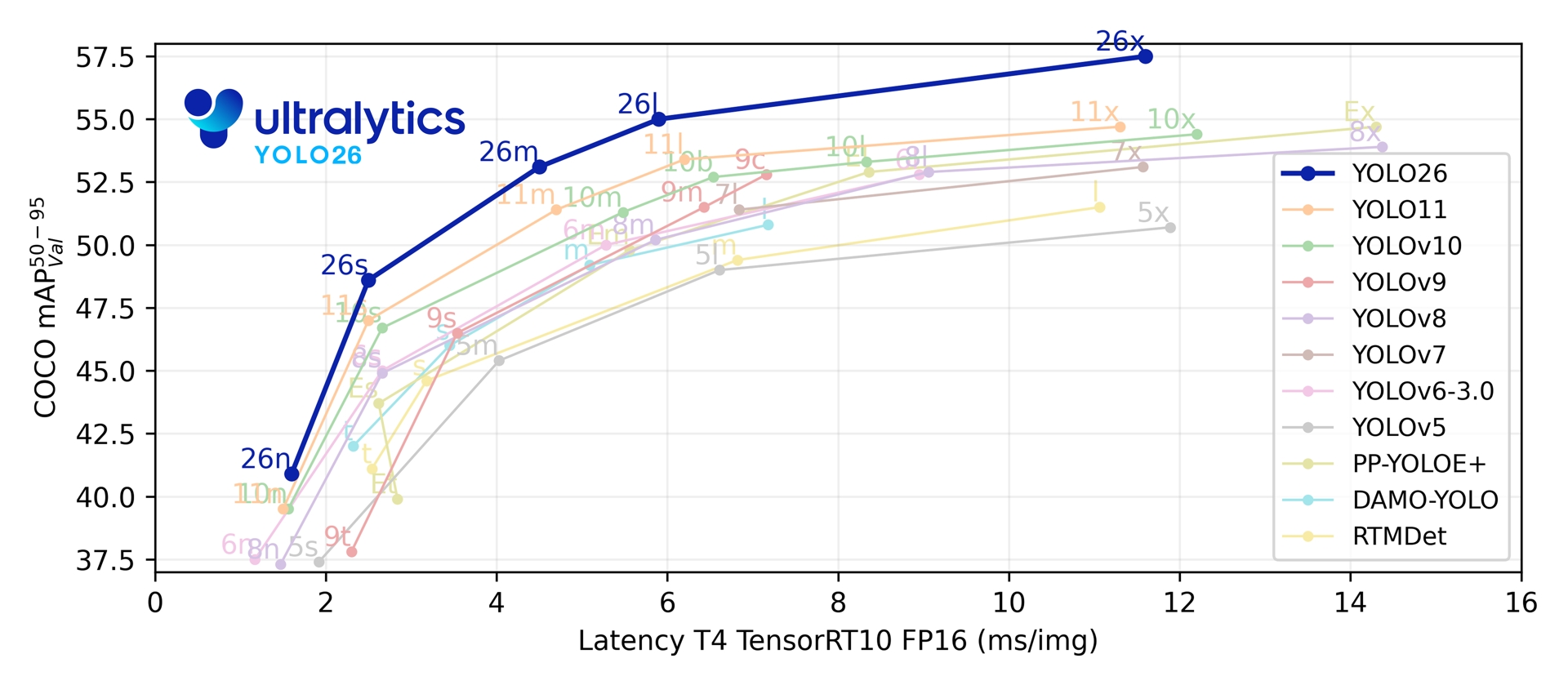} 
    \caption{Speed-Accuracy Trade-off on COCO val2017. The chart plots the Mean Average Precision (mAP 50-95) against inference latency (ms/img) on an NVIDIA T4 GPU (TensorRT10, FP16). The deep blue curve represents YOLO26, which forms a new Pareto front, consistently outperforming prior YOLO iterations (v5--v11) and state-of-the-art competitors by achieving higher accuracy at equivalent or lower latency.}
    \label{fig:benchmark_plot}
\end{figure}

\begin{figure}[h!]
    \centering
    \includegraphics[width=1.0\linewidth]{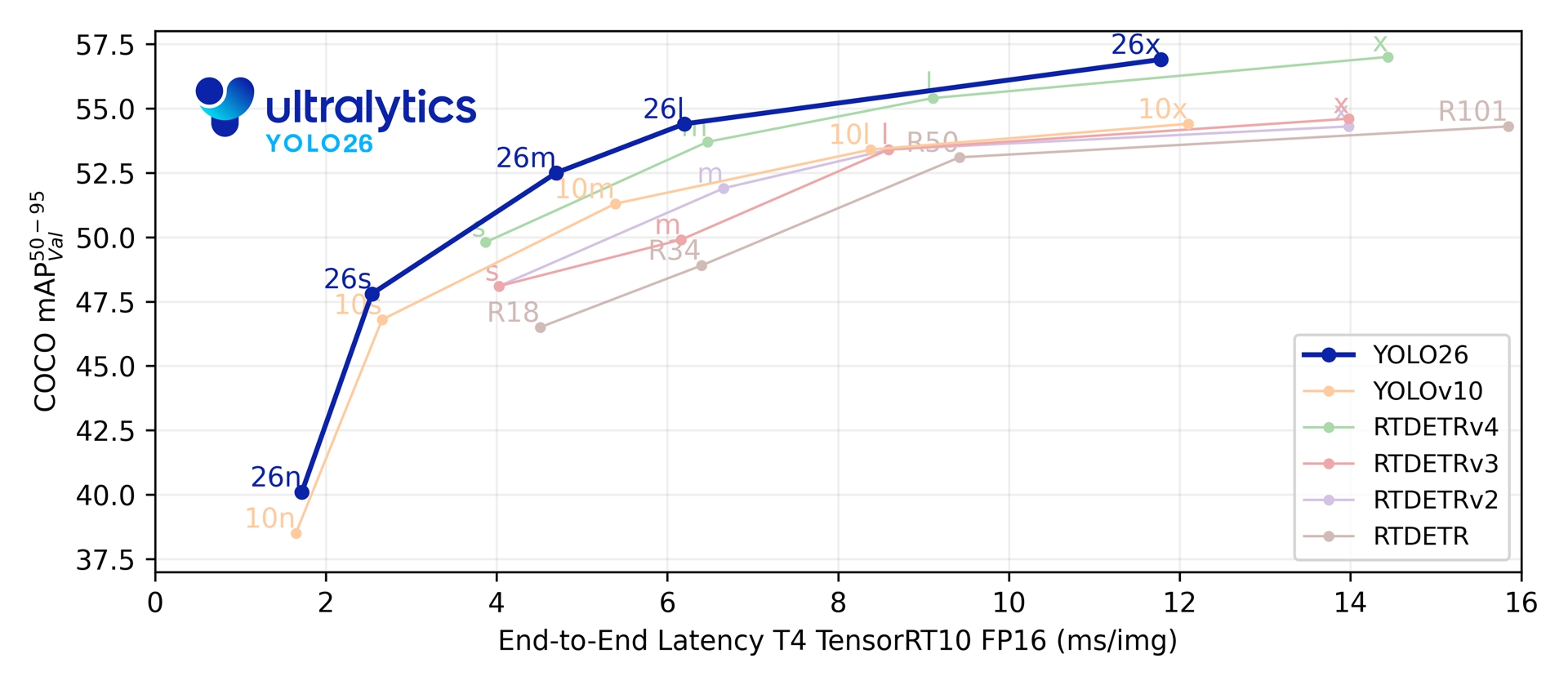} 
    \caption{Comparative Pareto frontier of YOLO26 against advanced end-to-end architectures on an NVIDIA T4 GPU (TensorRT10, FP16). YOLO26 strictly dominates recent competitors including YOLOv10 and the entire RT-DETR lineage (v2, v3, and v4) across all model scales.}
    \label{fig:detr_comparison_plot}
\end{figure}

\subsubsection{Analysis of Reported Performance}
As illustrated in the official benchmark data (Figures \ref{fig:benchmark_plot} and \ref{fig:detr_comparison_plot}) \cite{yolo26_ultralytics}, the performance landscape is definitively dominated by the YOLO26 family. 

\begin{itemize}
    \item \textbf{Absolute Pareto Dominance:} The reported metrics show that the YOLO26 curve resides strictly above and to the left of all other models. Figure \ref{fig:benchmark_plot} demonstrates its superiority over the legacy CNN-based YOLO lineage. More importantly, Figure \ref{fig:detr_comparison_plot} provides critical evidence that YOLO26 also outperforms state-of-the-art transformer-based detectors (including the latest RT-DETRv4 iterations). This proves that NMS-free CNN architectures can surpass heavy attention-based mechanisms in both speed and spatial reasoning.
    \item \textbf{Nano to Extra-Large Scaling:} The Ultralytics benchmarks highlight dominance across all model scales \cite{jocher2024yolo11}. The highly constrained nano variant (\textbf{26n}) is shown to achieve $>40$ mAP at a negligible latency of $\approx 1.5$ ms. At the high end, the extra-large model (\textbf{26x}) pushes the accuracy boundary to $\approx 57.5$ mAP while maintaining real-time performance ($\approx 11.5$ ms), surpassing both YOLO11x \cite{jocher2024yolo11} and massive DETR equivalents.
\end{itemize}

This empirical evidence provided by the developers confirms that the removal of NMS and the adoption of the end-to-end architecture have effectively unlocked raw throughput gains, cementing YOLO26's status as the fastest and most accurate detector currently documented.

\subsection{Contributions of This Article}
This study provides a comprehensive analysis of the YOLO26 architecture, evaluating its impact on the current state of real-time object detection. The primary contributions of this article are summarized as follows:

\begin{itemize}
    \item \textbf{Architectural Deconstruction:} This article presents a detailed breakdown of the Native End-to-End NMS-Free architecture, explaining the mathematical mechanisms that allow for the removal of non-differentiable post-processing.
    \item \textbf{Training Dynamics Analysis:} Novel optimization strategies—specifically MuSGD, STAL, and ProgLoss—are reviewed to elucidate how they enable stable convergence for lightweight, end-to-end backbones.
    \item \textbf{Comprehensive Benchmarking:} An exhaustive comparative study of YOLO26 is provided, evaluating its performance not only against prior YOLO lineages (v1--v13) but also against contemporary State-of-the-Art Transformer architectures (e.g., RT-DETR, DEIM, RF-DETR) to highlight its dominant speed-accuracy Pareto front.
    \item \textbf{Multi-Task \& Open-Vocabulary Evaluation:} The article analyzes the framework's unified multi-task extensions, specifically detailing the structural modifications of the YOLOE-26 open-vocabulary module and its capacity for zero-overhead promptable detection.
    \item \textbf{Impact Assessment:} The implications of resolving the "Export Gap" are discussed, providing an analysis of how deterministic latency and direct regression benefit safety-critical edge AI applications.
\end{itemize}

\subsection{Organization of the Paper}
The remainder of this article is structured as follows:
\textbf{Section \ref{sec:evolution}} traces the historical evolution of the YOLO lineage, setting the context for the current architectural shift.
\textbf{Section \ref{sec:architecture}} dissects the core innovations of YOLO26, including the NMS-Free pipeline, the DFL-free decoupled head, and the MuSGD training dynamics.
\textbf{Section \ref{sec:tasks}} details the model's unified multi-task capabilities, covering detection, segmentation, and pose estimation.
\textbf{Section \ref{sec:benchmarks}} presents the official performance benchmarks, featuring a comprehensive State-of-the-Art (SOTA) analysis.
\textbf{Section \ref{sec:impact}} analyzes the critical "Export Gap" challenge and how the architecture achieves deterministic latency on edge hardware.
\textbf{Section \ref{sec:future_work}} proposes future avenues for research, such as inherent explainability and spatiotemporal perception.
Finally, \textbf{Section \ref{sec:conclusion}} summarizes the contributions and potential impact of this work.

\vspace{-1mm}
\section{The Evolution of YOLO}
\label{sec:evolution}
\vspace{-2mm}
The YOLO (You Only Look Once) family has undergone a decade of rapid architectural evolution, transitioning from rigid grid-based detection to flexible, multi-task intelligence \cite{ali2024yolo,diwan2023object}. This progression can be categorized into three distinct eras: the Foundational Era (v1--v3), the Community Expansion Era (v4--v7), and the Modern Unified Era (v8--26). Each era is defined by a shift in how spatial features are extracted and how the final predictions are supervised.

\vspace{-1mm}
\subsection{The Foundational Era (2015--2018)}
\vspace{-1mm}
The original YOLOv1 \cite{redmon2016yolo} revolutionized object detection by reframing it as a single regression problem, sacrificing some localization accuracy for real-time speed. Subsequent iterations introduced anchor boxes in YOLOv2 \cite{redmon2017yolo9000} for improved recall and multi-scale feature pyramids in YOLOv3 \cite{redmon2018yolov3} to address the "small object problem," establishing the Darknet backbone as an industry standard. This era was characterized by the transition from fully connected layers to fully convolutional architectures, setting the precedent for global context reasoning in single-stage detectors.

\vspace{-1mm}
\subsection{The Community Expansion Era (2020--2022)}
\vspace{-1mm}
This period saw a diversification of the YOLO lineage, led by YOLOv4 \cite{bochkovskiy2020yolov4} and YOLOv5 \cite{jocher2020yolov5}, which introduced CSP (Cross-Stage Partial) connections and advanced "Bag-of-Freebies" augmentation techniques. This era marked the transition to production-ready frameworks, with variants like YOLOv6 \cite{li2022yolov6} and YOLOv7 \cite{wang2023yolov7} introducing re-parameterization and E-ELAN architectures to maximize hardware-specific compute utilization. By integrating mosaic augmentation and genetic anchor optimization, these models bridged the gap between academic research and industrial-scale deployment across diverse hardware targets.

\vspace{-1mm}
\subsection{The Modern Unified Era (2023--Present)}
\vspace{-1mm}
Starting with YOLOv8 \cite{jocher2023yolov8}, the focus shifted toward anchor-free, decoupled heads. This architectural modularity was further refined in YOLOv9 \cite{wang2024yolov9} through Programmable Gradient Information (PGI) and in YOLOv10 \cite{wang2024yolov10}, which introduced consistent dual-label assignment for NMS-free training. The lineage continued with YOLO11 \cite{jocher2024yolo11,chakrabortrydrone}, optimizing the C3k2 backbone for multi-task efficiency, and YOLOv12 \cite{yolov12_2025}, which integrated Area Attention ($A^2$) to provide transformer-level context at CNN speeds. Most recently, YOLOv13 \cite{lei2025yolov13} utilized hypergraph spatial modeling to improve relational reasoning in complex scenes. This transition reflects a broader movement toward eliminating manual heuristics in favor of end-to-end differentiable pipelines, paving the way for the edge-optimized strategies seen in the latest iterations.

A critical challenge identified in this era is the \textbf{"Export Gap"}---the performance drop observed when moving a model from a GPU-training environment to edge-inference hardware (NPUs/CPUs). Complex operators like Distribution Focal Loss (DFL) used in versions v8 through v13 \cite{jocher2023yolov8, wang2024yolov10, lei2025yolov13}, while accurate, often create latency bottlenecks on integer-arithmetic hardware. 

YOLO26 \cite{yolo26_ultralytics} represents the culmination of this lineage, departing from the complexity-heavy trends of v12 and v13 to prioritize edge-device latency. By removing the computational burden of DFL and adopting a native one-to-one prediction head, YOLO26 achieves deterministic inference times, rendering it highly effective for real-time deployment on low-power devices. These architectural shifts are summarized in Table \ref{tab:yolo_full}.

\begin{sidewaystable}[htbp]
\centering
\caption{Source-Safe Architectural Evolution of the YOLO Family (v1–26)}
\label{tab:yolo_full}
\small
\renewcommand{\arraystretch}{1.3} 
\setlength{\tabcolsep}{4pt}       

\begin{tabular}{|m{1.5cm}|m{2.1cm}|m{0.8cm}|m{1.6cm}|m{2.6cm}|m{0.7cm}|m{2.2cm}|m{1cm}|>{\raggedright\arraybackslash}m{7.5cm}|}
\hline
\textbf{Model} & \textbf{Backbone} & \textbf{Neck} & \textbf{Head} & \textbf{Task(s)} & \textbf{Anc-hors} & \textbf{Loss} & \textbf{Post-Proc.} & \textbf{Key Innovations \& Contributions} \\ \hline

YOLOv1 \newline (2015) & Darknet-24 & None & Coupled & Object Detection & No & SSE (Sum) & NMS &
Unified single-stage regression framework enabling real-time object detection. \\ \hline

YOLOv2 \newline (2016) & Darknet-19 & Pass-thro-ugh & Coupled & Object Detection & Yes & Sum-Squared Error (SSE) & NMS &
Introduced anchor boxes, batch normalization, and the passthrough layer for improved recall and small-object detection. \\ \hline

YOLOv3 \newline (2018) & Darknet-53 & Multi-Scale & Coupled & Object Detection & Yes & BCE + SSE & NMS &
Multi-scale feature prediction strategy for enhanced small-object localization. \\ \hline

YOLOv4 \newline (2020) & CSPDarknet53 & PAN & Coupled & Object Detection & Yes & CIoU + BCE & NMS &
CSP-integrated augmentation for optimal speed–accuracy trade-off. \\ \hline

YOLOv5 \newline (2020) & CSPDarknet & PAN & Coupled & Object Detection & Yes & GIoU/CIoU + BCE & NMS &
PyTorch-based modular design with automatic anchor optimization for easy deployment. \\ \hline

YOLOv6 \newline (2022) & EfficientRep & PAN & Decoupled & Object Detection & Yes & SIoU / Varifocal & NMS &
Re-parameterized convolution for high-throughput industrial inference efficiency. \\ \hline

YOLOv7 \newline (2022) & E-ELAN & CSP-PAN & Lead + Auxiliary & Object Detection & Yes & CIoU + BCE & NMS &
Introduced E-ELAN, deep supervision and OTA assignment for better accuracy and efficiency. \\ \hline

YOLOv8 \newline (2023) & C2f & PAN & Decoupled & Obj. Det., Seg., Pose Est. & No & BCE + CIoU + DFL & NMS &
Anchor-free decoupled head enabling a unified multi-task detection framework. \\ \hline

YOLOv9 \newline (2024) & GELAN & PAN & Decoupled & Object Detection & No & BCE + CIoU + DFL & NMS &
Programmable Gradient Information \& GELAN to overcome info. bottleneck in deep networks. \\ \hline

YOLOv10 \newline (2024) & GELAN & PAN & Decoupled & Object Detection & No & BCE + CIoU + DFL & NMS-Free &
NMS-free inference via Dual-Label Assignment; integrates Partial Self-Attention into GELAN. \\ \hline

YOLOv11 \newline (2024) & C3k2 & PAN & Decoupled & Obj. Det., Seg., Pose Est. & No & BCE + CIoU + DFL & NMS &
C2PSA-based feature refinement; still uses standard NMS for post-processing. \\ \hline

YOLOv12 \newline (2025) & Flash Backbone + Area Attention & PAN & Decoupled & Object Detection, Segmentation & No & BCE + CIoU + DFL & NMS &
Uses Area Attention ($A^2$) for long-range dependency capture while keeping computation efficient; improves multi-task performance. \\ \hline

YOLOv13 \newline (2025) & Hyper-Net & PAN & Decoupled & Object Detection, Segmentation, Pose Estimation & No & BCE + CIoU + DFL & NMS &
Third-party release by iMoonLab; Hypergraph spatial modeling for relational reasoning and complex scene understanding. \\ \hline

YOLO26 \newline (2026) & CSP-Muon (Edge-Optimized CNN) & PAN & Decoupled (1-to-1) & Object Detection, Segmentation, Pose Estimation, OBB & No & STAL + ProgLoss & NMS-Free &
Edge-optimized, DFL-free learning with one-to-one label assignment; native NMS-free head for low-latency deployment; optimized for CPU and Edge exportability. \\ \hline

\end{tabular}
\end{sidewaystable}

\section{Architecture and Methodology of YOLO26}
\label{sec:architecture}
The architectural philosophy of YOLO26 diverts from the recent trend of increasing parameter complexity (as seen in v10 and v11)\cite{jocher2024yolo11} to focus on \textit{computational density} and \textit{deterministic latency}. This is achieved by restructuring the inference pipeline to remove heuristic bottlenecks and adopting optimization strategies traditionally reserved for LLMs, such as MuSGD.

\subsection{Native End-to-End NMS-Free Architecture}
\label{subsec:backbone}
Traditional object detectors rely on Non-Maximum Suppression (NMS) as a distinct post-processing step to filter redundant bounding boxes. NMS functions by iteratively selecting the proposal with the highest confidence score ($S_{max}$) and suppressing all other overlapping boxes ($b_i$) whose Intersection over Union (IoU) with $S_{max}$ exceeds a predefined threshold ($N_t$). This process can be formally defined as \cite{bodla2017softnms}:

\begin{equation}
s_i = \begin{cases} 
s_i, & \text{if } IoU(M, b_i) < N_t \\ 
0, & \text{if } IoU(M, b_i) \geq N_t 
\end{cases}
\label{eq:nms}
\end{equation}

\noindent where $M$ is the current maximum confidence box and $s_i$ is the updated score. This heuristic is inherently sequential, creating a latency bottleneck that varies depending on scene density (i.e., the number of detected objects).

\begin{figure}[h!]
    \centering
    \includegraphics[width=0.95\linewidth, height=6cm]{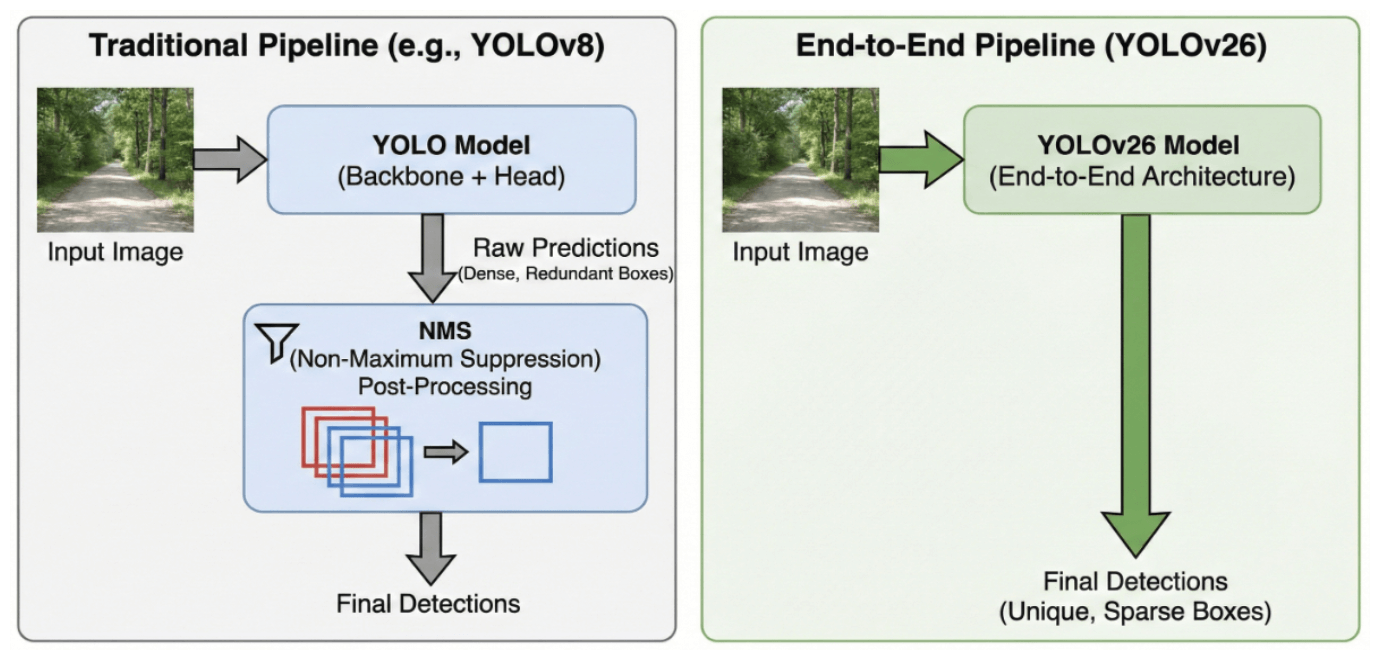} 
    \caption{Comparison of Inference Pipelines. (Left) Traditional YOLOv8 pipeline requiring sequential NMS post-processing. (Right) YOLO26 End-to-End pipeline where the model directly outputs unique predictions, reducing latency and complexity.}
    \label{fig:nms_comparison}
\end{figure}
YOLO26 fundamentally alters this pipeline through a \textbf{Native End-to-End Architecture}. By redesigning the prediction head to support \textbf{one-to-one label assignment} \cite{wang2024yolov10}, the model learns to output a single, definitive box per object instance during training. This architectural shift eliminates the need for Eq. \ref{eq:nms} entirely, transforming inference from a multi-stage filtering operation into a direct, deterministic mapping of input to output (see Fig. \ref{fig:nms_comparison}). The result is a lighter, streamlined execution graph that is easier to deploy and achieves constant-time latency regardless of object count \cite{lyu2022rtmdet}.

\textbf{Performance Impact: }The removal of the NMS operator yields significant latency reductions, particularly on non-GPU hardware where sequential operations create bottlenecks. By transitioning to this end-to-end paradigm, Ultralytics reports that YOLO26 achieves an inference speedup of approximately \textbf{43\% on CPU targets} compared to standard NMS-based baselines \cite{yolo26_ultralytics}. This constant-time inference is critical for safety-critical applications, such as autonomous driving or medical monitoring, where deterministic response times are required regardless of scene complexity.

\subsection{Regression-Centric Decoupled Head (DFL-Free)}
\label{subsec:head}
Recent YOLO iterations (v8--v11 \cite{jocher2024yolo11}) adopted Distribution Focal Loss (DFL) \cite{li2020generalized} to model bounding box coordinates as general distributions rather than deterministic values. While DFL improves localization accuracy by accounting for uncertainty at object boundaries, it introduces a significant computational overhead: the necessity of performing Softmax operations over discretized bins for every coordinate prediction. On specialized edge hardware (NPUs and DSPs), these Softmax layers are notoriously difficult to quantize and often become the primary latency bottleneck \cite{gholami2021survey}.

\textbf{Quantification of Softmax Overhead: }In a DFL-based head, estimating a single coordinate $y$ requires integrating over a discretized probability distribution (typically 16 bins). This forces the inference engine to compute a weighted Softmax summation for every bounding box parameter:

\begin{equation}
\hat{y}_{DFL} = \sum_{i=0}^{n} i \cdot \text{Softmax}(w_i) = \sum_{i=0}^{n} i \cdot \frac{e^{w_i}}{\sum_{j=0}^{n} e^{w_j}}
\label{eq:dfl_cost}
\end{equation}

\noindent This operation involves repeated exponential ($e^x$) and division calculations, which are computationally expensive on integer-arithmetic edge accelerators \cite{sharma2018bitfusion}.

\begin{figure}[h!]
    \centering
    \includegraphics[width=0.96\linewidth, height= 7cm]{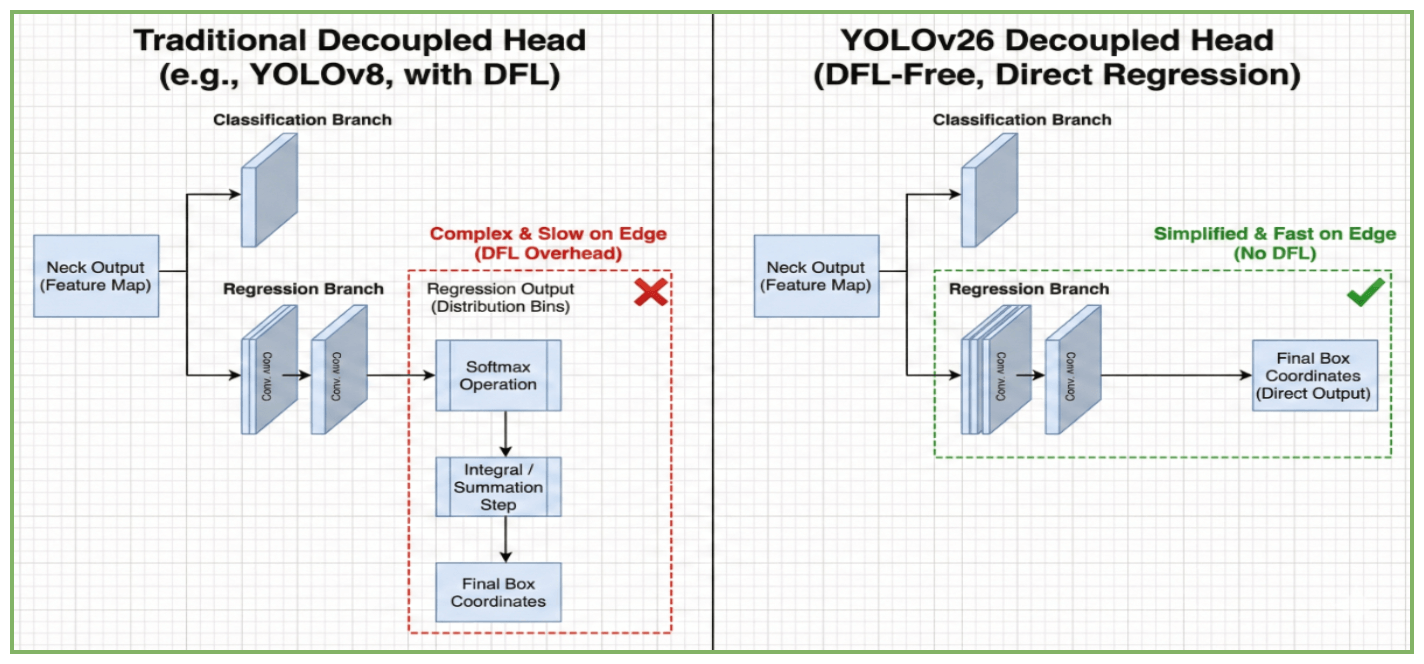} 
    \caption{Architectural comparison of the prediction heads. (Left) Traditional Decoupled Head utilizing Distribution Focal Loss (DFL), (Right) YOLO26 Decoupled Head employing the streamlined Direct Regression strategy, eliminating DFL overhead for optimized edge inference.}
    \label{fig:dfl_comparison}
\end{figure}

YOLO26 reverts to a \textbf{Direct Regression Strategy}, removing this module entirely (see Fig. \ref{fig:dfl_comparison}). This architectural rollback is motivated by the "Export Gap"—the discrepancy between theoretical FLOPs and actual inference speed on deployed hardware \cite{lyu2022rtmdet}. By eliminating the integral representation of Eq. \ref{eq:dfl_cost}, the decoding phase is simplified to a direct linear mapping:

\begin{equation}
\hat{y}_{v26} = \mathcal{F}_{reg}(x) \in \mathbb{R}
\label{eq:direct_reg}
\end{equation}

To maintain high precision without the distributional benefits of DFL, YOLO26 employs a refined \textbf{Decoupled Head} structure inspired by YOLOX \cite{ge2021yolox}. As illustrated in standard topologies, the head separates feature extraction into two distinct branches:
\begin{equation}
Head(x) = \{ \mathcal{F}_{cls}(x), \mathcal{F}_{reg}(x) \}
\end{equation}
\noindent where $\mathcal{F}_{cls}$ predicts class probabilities and $\mathcal{F}_{reg}$ predicts box regression parameters directly. This separation ensures that the removal of DFL does not degrade classification performance \cite{ge2021yolox}, while the regression branch is optimized via the new STAL and ProgLoss functions to recover the localization precision lost by discarding the distributional prior.

\subsection{Advanced Training Dynamics: MuSGD, STAL, and ProgLoss}
\label{subsec:training}
The removal of the Distribution Focal Loss (DFL) module and the transition to an end-to-end architecture necessitate a more robust training strategy to prevent gradient collapse. YOLO26 addresses this through a triad of optimization and supervision innovations.

\subsubsection{MuSGD Optimizer}
To ensure convergence stability within the new architecture, Ultralytics reports that YOLO26 introduces \textbf{MuSGD (Momentum-Unified Stochastic Gradient Descent)}, a novel hybrid optimizer that fuses the properties of standard SGD with the \textit{Muon} optimizer. Explicitly inspired by the training dynamics of Moonshot AI's \textit{Kimi K2} large language model, MuSGD represents a strategic transfer of advanced optimization methods from the NLP domain into computer vision \cite{keller2024muon}.

\textbf{The Muon Component: }The core innovation of MuSGD lies in its integration of the Muon optimizer \cite{liu2025muon}. Unlike element-wise optimizers (e.g., AdamW), Muon performs \textbf{matrix orthogonalization}, updating the entire weight matrix to be orthogonal to its current state. This maximizes update efficiency along the most impactful directions while restraining the spectral norm \cite{jordan2024ortho}.

\begin{figure}[h!]
    \centering
    \includegraphics[width=0.9\linewidth]{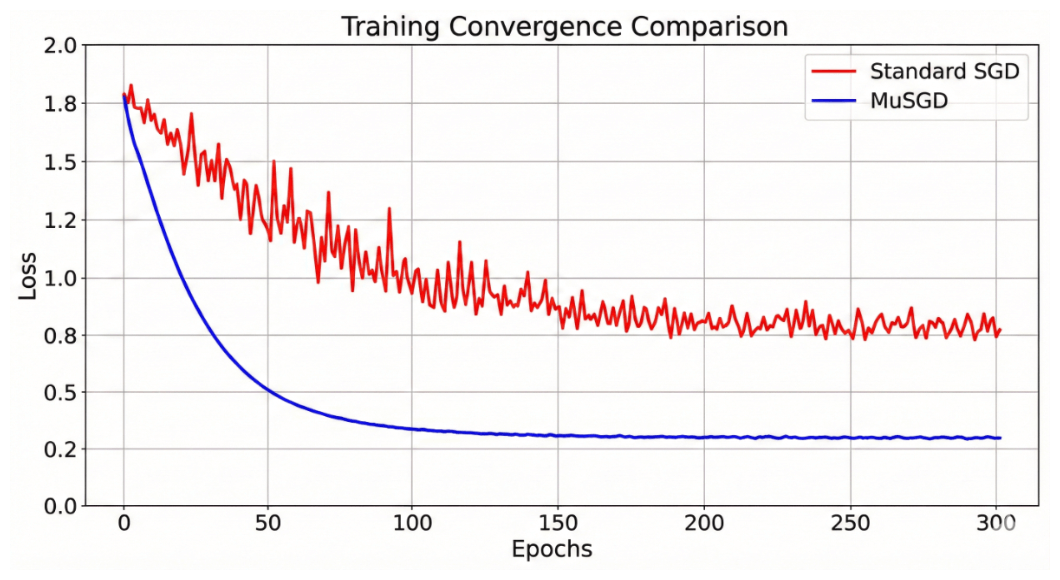} 
    \caption{\textbf{Conceptual visualization} of the expected optimization dynamics. The MuSGD strategy (Blue) is designed to mitigate the gradient variance observed in standard SGD (Red), theoretically allowing for a steeper learning trajectory without warm-up.}
    \label{fig:musgd_convergence}
\end{figure}

\textbf{Mathematical Formulation: }MuSGD combines this orthogonal scaling with the stability of classical SGD. First, we define the standard momentum buffer $v_t$ used in Stochastic Gradient Descent:

\begin{equation}
\vspace{-2mm}
v_{t+1} = \beta \cdot v_t + g_t
\label{eq:sgd_momentum}
\end{equation}

\noindent where $g_t$ is the gradient and $\beta$ is the momentum coefficient. MuSGD then modifies the final weight update by injecting the Newton-Schulz orthogonalization into this trajectory:

\begin{equation}
\theta_{t+1} = \theta_t - \eta \cdot \left( \alpha \cdot v_{t+1} + (1-\alpha) \cdot \text{NewtonSchulz}(g_t) \right)
\label{eq:musgd}
\end{equation}

\noindent where $\text{NewtonSchulz}(g_t)$ effectively "whitens" the gradient matrix using an iterative refinement process \cite{higham1986newton}. This hybrid approach mitigates the variance of pure SGD while avoiding the instability of pure orthogonal updates in the early epochs (see Fig. \ref{fig:musgd_convergence}). By enabling the simplified end-to-end backbone to learn robust features without the need for complex warm-up schedules, MuSGD reduces the total training time required to reach convergence.

\subsubsection{Small-Target-Aware Label Assignment (STAL)}
To address the "small object vanishing" problem inherent in edge-optimized models \cite{kisantal2019augmentation}, YOLO26 implements \textbf{Small-Target-Aware Label Assignment (STAL)}. 
Standard assignment strategies typically rely on a fixed Intersection-over-Union (IoU) threshold (e.g., $\tau=0.5$). While effective for large objects, this rigid threshold is detrimental to small targets (occupying $<1\%$ of the image area), where even well-centered anchors yield mathematically low IoU scores due to pixel-level discretization errors and the sensitivity of the IoU metric to small spatial shifts \cite{rezatofighi2019giou}.

\begin{figure}[h!]
    \centering
    \includegraphics[width=1.0\linewidth]{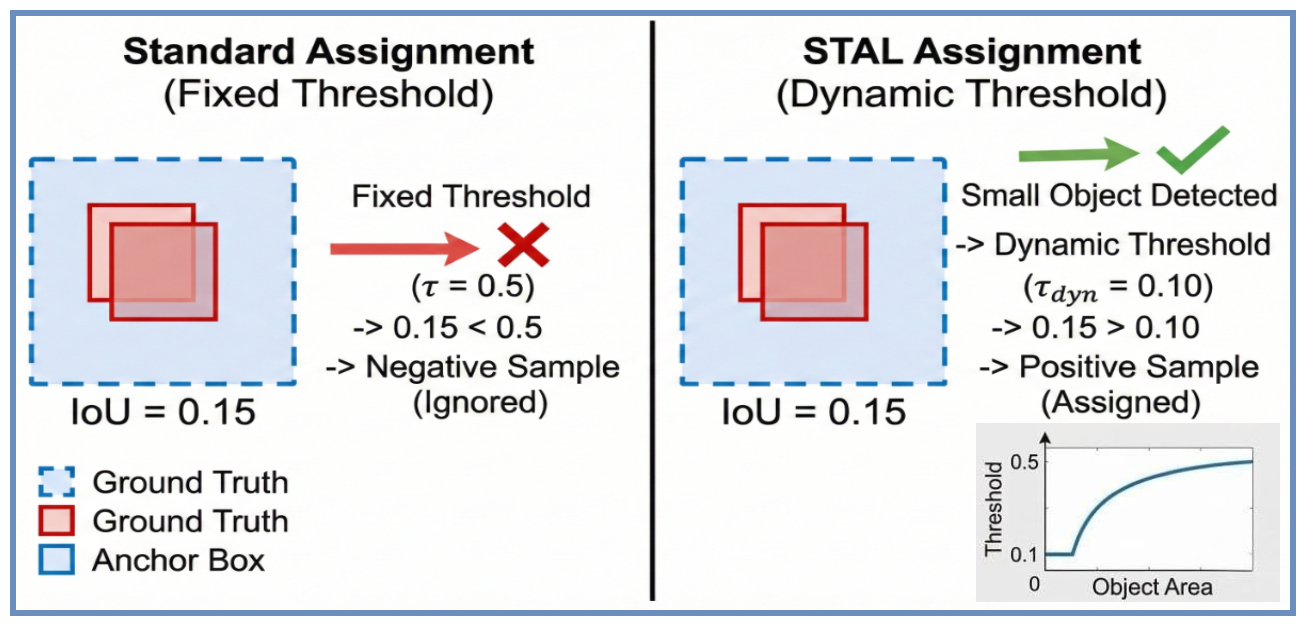} 
    \caption{Mechanism of Small-Target-Aware Label Assignment (STAL). (Left) Standard assignment ignores the small target because its IoU (0.15) is below the fixed threshold (0.5). (Right) STAL detects the small area ratio and dynamically lowers the threshold to 0.10, successfully assigning the anchor as a positive sample for training.}
    \label{fig:stal_viz}
\end{figure}

STAL resolves this by replacing the static threshold with a dynamic variable that adapts to the object's scale, drawing inspiration from Task Alignment Learning (TAL) \cite{feng2021tood}. As defined in Eq. \ref{eq:stal}, the matching threshold $\tau$ relaxes as the relative object size decreases:

\begin{equation}
\tau_{dynamic} = \tau_{base} \cdot \left(1 - \alpha \cdot e^{-\frac{\text{Area}_{obj}}{\text{Area}_{img}}}\right)
\label{eq:stal}
\end{equation}

\noindent where $\alpha$ controls the decay rate. For a tiny object, the exponential term approaches 1, significantly lowering $\tau_{dynamic}$ and allowing anchors with lower physical overlap to still be assigned as positive samples. This acts as a "magnifying glass" for supervisory signals, ensuring that tiny or occluded objects—common in drone imagery and medical scans—receive adequate gradient contribution \cite{chen2022tiny} (see Fig. \ref{fig:stal_viz}).

\subsubsection{Progressive Loss Balancing (ProgLoss)}
To further stabilize the training of the end-to-end architecture, YOLO26 employs \textbf{ProgLoss}, a dynamic loss weighting strategy. In standard detectors \cite{jocher2023yolov8, lin2017focal}, the ratio between classification loss ($L_{cls}$) and bounding box regression loss ($L_{box}$) is typically fixed. However, this static balance is suboptimal for end-to-end learning, where the network must simultaneously learn feature discrimination and precise localization without the geometric guidance of anchor priors \cite{hossain2019curriculum}.

\begin{figure}[h!]
    \centering
    \includegraphics[width=0.9\linewidth]{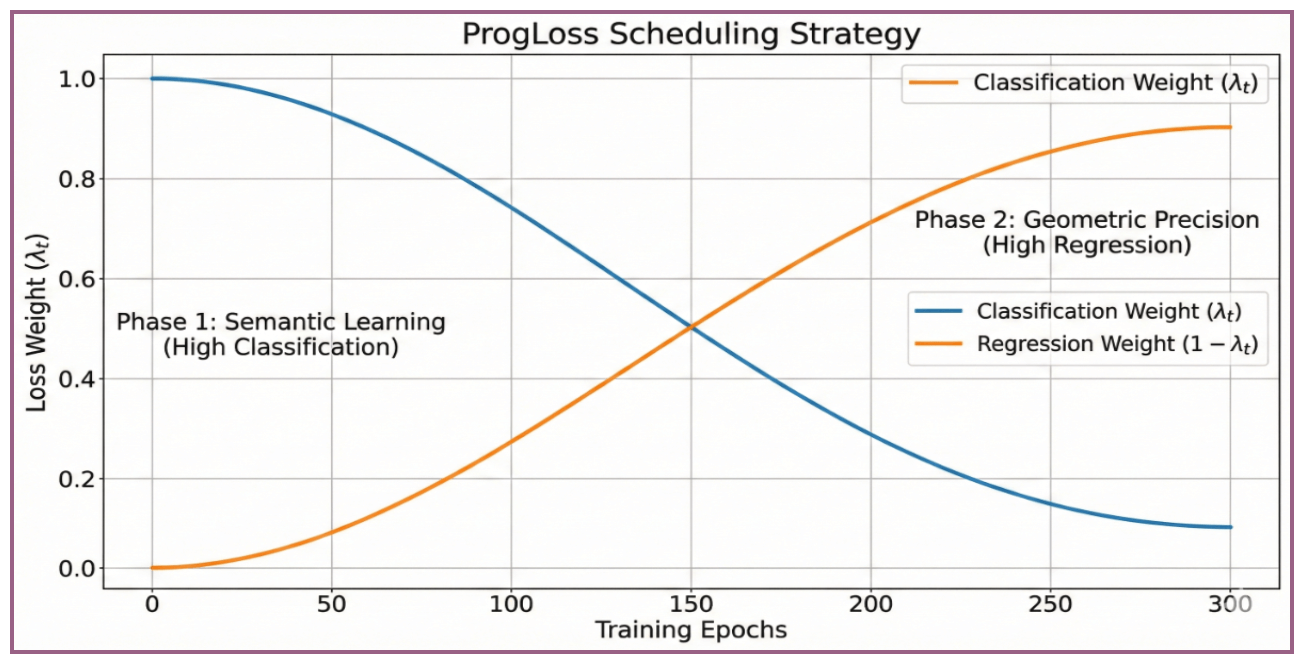} 
    \caption{\textbf{Conceptual visualization} of the proposed ProgLoss scheduling strategy. The chart illustrates the intended dynamic balancing, where the classification weight ($\lambda_t$, blue) dominates the early "Semantic Learning" phase to stabilize training, and the regression weight (orange) progressively increases to prioritize "Geometric Precision" in the final epochs.}
    \label{fig:progloss}
\end{figure}

ProgLoss addresses this by introducing a time-dependent modulation coefficient ($\lambda_t$). As shown in Eq. \ref{eq:progloss} and illustrated in Fig. \ref{fig:progloss}, the total loss evolves across training epochs $t$.

\begin{equation}
L_{total}(t) = \lambda_t \cdot L_{cls} + (1 - \lambda_t) \cdot L_{box}
\label{eq:progloss}
\end{equation}

\noindent where $\lambda_t$ follows a monotonically decreasing schedule, such as cosine decay \cite{loshchilov2016sgdr}. This strategy ensures a smooth transition between semantic grounding and geometric refinement.

\begin{itemize}
    \item \textbf{Early Phase (High $\lambda_t$):} As seen in the blue region of Fig. \ref{fig:progloss}, the gradient is initially dominated by $L_{cls}$. This prioritizes the learning of high-level semantic features to stabilize the backbone and establish object existence \cite{kendall2018multi}.
    \item \textbf{Late Phase (Low $\lambda_t$):} As training progresses (orange region), the focus shifts to $L_{box}$, allowing the model to fine-tune geometric boundaries. This prevents "easy negatives" from dominating the gradient in the final stages, ensuring high-precision localization despite the removal of DFL.
\end{itemize}

\section{Multi-Task Capabilities of YOLO26}
\label{sec:tasks}
\begin{figure}[h!]
    \centering
    \includegraphics[width=0.97\linewidth, height=10cm]{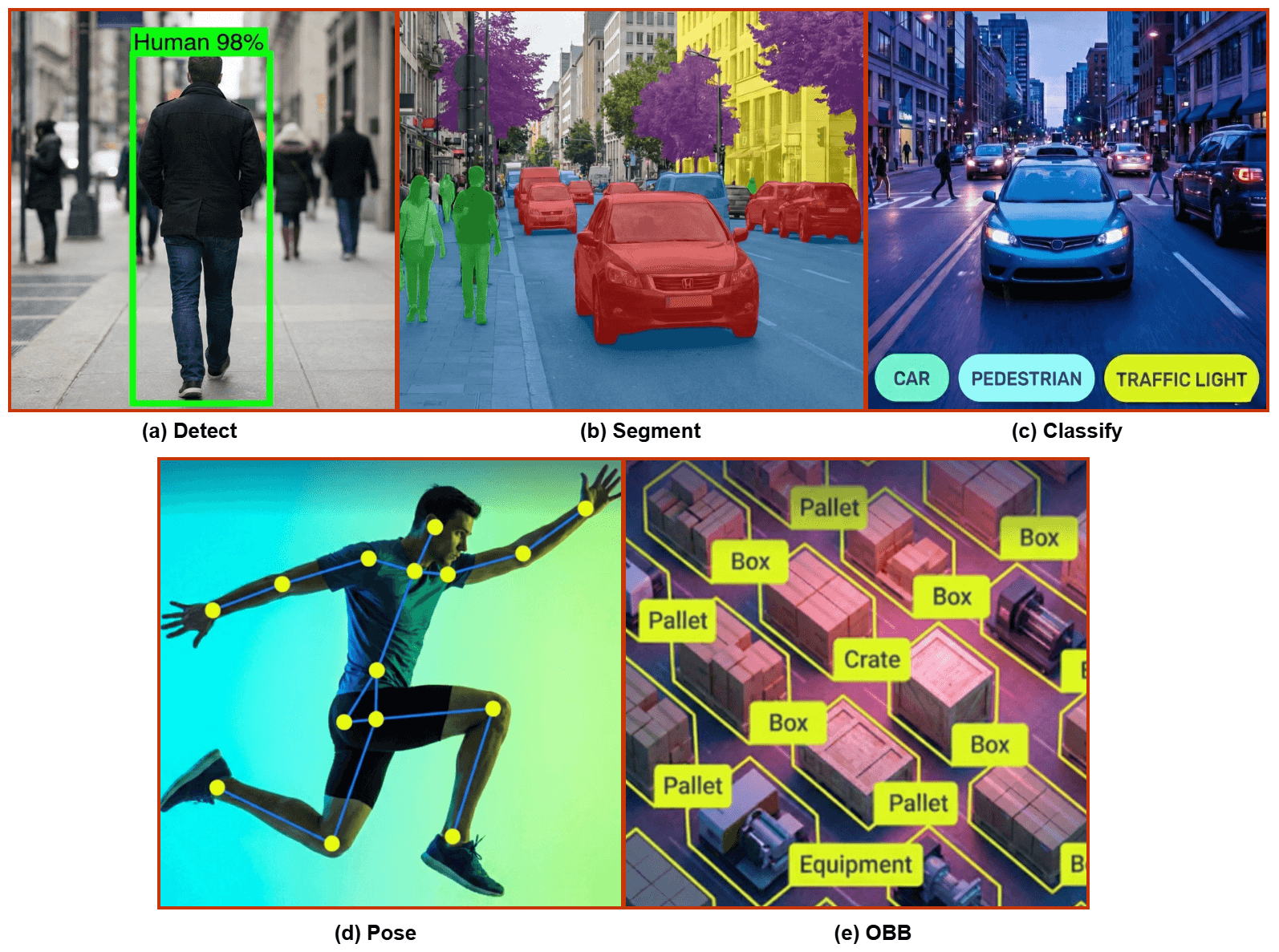} 
    \caption{Unified multi-task execution in YOLO26, demonstrating (a) Detection, (b) Segmentation, (c) Classification, (d) Pose Estimation, and (e) Oriented Bounding Box (OBB) detection}
    \label{fig:yolo_tasks}
\end{figure}

\vspace{-2mm}
YOLO26 functions as a unified model family, providing end-to-end support for a diverse range of computer vision tasks \cite{yolo26_ultralytics}. Each architectural variant, from Nano (n) to Extra-Large (x), is natively compatible with specialized prediction heads designed for distinct spatial and semantic reasoning challenges. As illustrated in Figure \ref{fig:yolo_tasks}, the framework moves beyond simple object detection to facilitate a comprehensive suite of analytical capabilities within a single, optimized inference pipeline.

Beyond visual representation, the technical execution of these tasks is governed by specialized output structures and loss functions tailored for edge efficiency. Table \ref{tab:tasks_summary} provides a comparative summary of the head outputs and coordinate formats employed by the YOLO26 family to maintain architectural consistency across varied domains. This multi-task framework leverages the unified backbone and the aforementioned ProgLoss scheduling to ensure that the transition from standard bounding boxes to more complex geometries—such as keypoints and oriented boxes—does not incur a significant latency penalty.

\begin{table}[h!]
\centering
\small
\renewcommand{\arraystretch}{1.2}
\caption{Summary of YOLO26 Multi-Task Support and Task-Specific Head Designs}
\label{tab:tasks_summary}
\begin{tabular}{|l|p{2.5cm}|l|p{4.8cm}|}
\hline
\textbf{Task} & \textbf{Head Output} & \textbf{Coordinate Format} & \textbf{Head Mechanism / Objective} \\ \hline
Object Detection & Class + Box & $(x_c, y_c, w, h)$ & NMS-Free Detection, STAL Loss \\ \hline
Instance Segmentation & Class + Box + Mask & $(x_c, y_c, w, h) + \text{Mask}_{pix}$ & Prototype Mask Head, ProgLoss \\ \hline
Classification & Class Label & $\text{None}$ (Global Label) & Global Pooling, Linear Classification Head \\ \hline
Pose Estimation & Class + Box + Keypoints & $(x_i, y_i, v_i)_{i=1}^{17}$ & OKS-based Keypoint Optimization \\ \hline
Oriented Detection & Class + Rotated Box & $(x_c, y_c, w, h, \theta)$ & Rotated IoU / Angle-Aware Loss \\ \hline
Open-Vocabulary & Text + Box & $(x_c, y_c, w, h) + \text{Embed}_{txt}$ & Vision--Language Embedding Alignment \\ \hline
\end{tabular}
\end{table}

\subsection{Object Detection}
\label{subsec:detection}
The primary objective of YOLO26 is the identification and localization of discrete object instances via axis-aligned bounding boxes, as demonstrated in Figure \ref{fig:yolo_tasks}(a). While this remains the foundational task of the YOLO series, YOLO26 optimizes the detection pipeline by leveraging the native end-to-end architecture discussed in Section \ref{subsec:backbone}. By utilizing the one-to-one label assignment strategy, the model achieves a 43\% reduction in CPU latency \cite{yolo26_ultralytics}, a critical factor for real-time medical monitoring and edge-tier surveillance. Beyond raw speed, the removal of the non-differentiable NMS operator ensures that the detection process is fully deterministic. This predictability is vital for the fidelity of explainability methods, providing a direct, transparent path from pixel input to final box output.

The detection of minute features is further bolstered by the STAL mechanism described in Eq. \ref{eq:stal}. In practical applications, such as the analysis of micro-anomalies in histopathological datasets, STAL prevents the "vanishing gradient" effect typically associated with small targets. This allows YOLO26 to maintain high recall for objects occupying less than 1\% of the image area, ensuring that the streamlined, DFL-free regression head remains precise across all object scales.

\subsection{Instance Segmentation}
\label{subsec:segmentation}
Instance segmentation in YOLO26 represents a critical shift from regional localization to pixel-wise classification, as illustrated in Figure \ref{fig:yolo_tasks}(b). By integrating a mask-prediction branch alongside the decoupled head, the model facilitates precise shape extraction for individual objects. As summarized in Table \ref{tab:tasks_summary}, the head output for this task includes both bounding box coordinates and a pixel-level mask ($\text{Mask}_{pix}$), which is vital for medical diagnostics where the exact area of a pathology provides more value than a simple coordinate box.

A novel refinement in YOLO26-seg is the use of \textbf{Boundary-Aware Supervision}, supported by the ProgLoss scheduling in Eq. \ref{eq:progloss}. Because the model is DFL-free, it avoids the discretization errors that often blur object edges on edge hardware. Instead, the late-stage regression focus of ProgLoss acts as a "contour polisher," ensuring that the masks remain sharp even for small or overlapping targets. By leveraging the MuSGD optimizer's ability to maintain stable spectral norms, the segmentation branch achieves higher feature resolution with fewer parameters, leading to the previously noted speedup on CPU and NPU targets. This ensures that high-fidelity segmentation is no longer restricted to high-end GPUs but is fully exportable to real-time edge environments \cite{yolo26_ultralytics}.

\subsection{Image Classification}
\label{subsec:classification}
Image classification within the YOLO26 ecosystem represents the most computationally efficient task, as it bypasses the requirement for spatial regression or mask generation, as shown in Figure \ref{fig:yolo_tasks}(c). By analyzing the input holistically, the classification head utilizes Global Average Pooling (GAP) to condense the high-level feature maps from the backbone into a single vector, which is then mapped to categorical probabilities \cite{lin2013network}. This architecture prioritizes overarching visual patterns over specific coordinate-based boundaries, as summarized in Table \ref{tab:tasks_summary}.

The YOLO26-cls variant leverages the streamlined CSP-based backbone to achieve minimal inference latency, making it ideal for the initial categorization of large-scale medical or environmental datasets where the presence of a pathology or object is the primary metric \cite{yolo26_ultralytics}. Furthermore, the integration of \textbf{ProgLoss} scheduling (Eq. \ref{eq:progloss}) ensures that the classification head achieves stable convergence on complex, multi-class datasets. By focusing on semantic grounding during the early training phase, the model establishes robust global representations that are less sensitive to spatial noise or object occlusion compared to purely regional detectors \cite{he2016deep}.

\subsection{Pose Estimation}
\label{subsec:pose}
Pose estimation in YOLO26 extends spatial reasoning to the localization of 17 anatomical landmarks, as visualized in Figure \ref{fig:yolo_tasks}(d). This task tracks the orientation and movement of joints by outputting a triplet format $(x_i, y_i, v_i)$ for each keypoint. The specific anatomical indices for the default COCO-based mapping \cite{lin2014microsoft}.  are detailed in Table \ref{tab:keypoint_mapping}.

\begin{table}[h!]
\centering
\small 
\caption{YOLO26 Default 17-Keypoint Mapping}
\label{tab:keypoint_mapping}
\begin{tabular}{|cl|cl|cl|}
\hline
\textbf{Idx} & \textbf{Joint} & \textbf{Idx} & \textbf{Joint} & \textbf{Idx} & \textbf{Joint} \\ \hline
0 & Nose           & 6  & Right Shoulder & 12 & Right Hip    \\
1 & Left Eye       & 7  & Left Elbow     & 13 & Left Knee    \\
2 & Right Eye      & 8  & Right Elbow    & 14 & Right Knee   \\
3 & Left Ear       & 9  & Left Wrist     & 15 & Left Ankle   \\
4 & Right Ear      & 10 & Right Wrist    & 16 & Right Ankle  \\
5 & Left Shoulder  & 11 & Left Hip       &    &              \\ \hline
\end{tabular}
\end{table}

Accuracy is governed by the \textbf{Object Keypoint Similarity (OKS)}, which normalizes Euclidean distance $d_i$ against the object scale $s$ and a per-joint falloff constant $\kappa_i$:

\begin{equation}
OKS = \frac{\sum_{i} \exp(-d_i^2 / 2s^2 \kappa_i^2) \delta(v_i > 0)}{\sum_{i} \delta(v_i > 0)}
\label{eq:oks}
\end{equation}

To maintain precision in the absence of DFL, YOLOv26-pose utilizes \textbf{Residual Log-Likelihood Estimation (RLE)} \cite{li2021human}. By modeling spatial uncertainty rather than a fixed distribution, RLE allows the model to reason through occlusions. Combined with the \textbf{MuSGD} optimizer, this ensures high-fidelity keypoint regression with deterministic latency on edge hardware.

\subsection{Oriented Object Detection (OBB)}
\label{subsec:obb}
Oriented Object Detection (OBB) in YOLO26 introduces a rotational parameter ($\theta$) to precisely localize skewed targets, as illustrated in Figure \ref{fig:yolo_tasks}(e). By utilizing the normalized \texttt{xywhr} format detailed in Table \ref{tab:tasks_summary}, the model eliminates the background noise typical of axis-aligned boxes in aerial and industrial domains \cite{ding2019learning}. To resolve boundary discontinuity errors inherent in angular regression, the architecture employs a specialized \textbf{Angle Loss} that maintains geometric consistency even for near-square objects \cite{yang2021r3det}.

This task leverages the \textbf{Direct Regression} strategy and the \textbf{MuSGD} optimizer to achieve high angular precision without the computational overhead of distributional focal loss. When deployed on edge-tier hardware like UAVs, the \textbf{NMS-free} head enables deterministic latency in dense environments, such as shipping ports. These optimizations result in a 43\% inference speedup compared to traditional heuristic-based rotational NMS baselines \cite{yolo26_ultralytics}, ensuring real-time performance on resource-constrained devices.

\subsection{Open-Vocabulary Detection and Segmentation (YOLOE-26)}
\label{subsec:openvoc}
\begin{figure}[h!]
    \centering
    \includegraphics[width=0.99\linewidth]{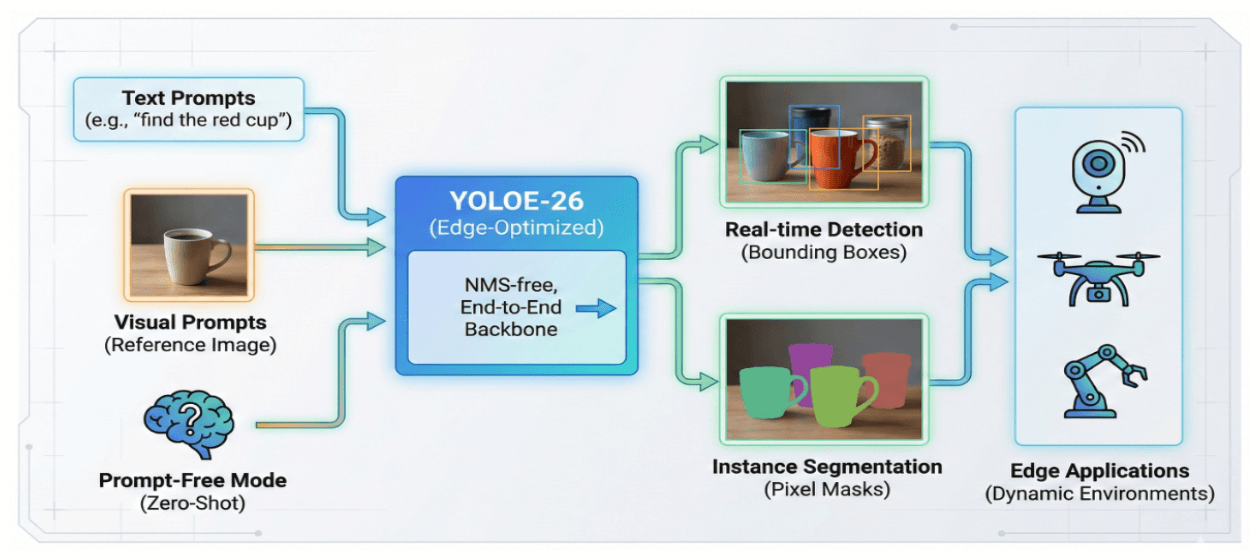} 
    \caption{Conceptual overview of the YOLOE-26 open-vocabulary architecture illustrating multi-modal input processing for real-time edge detection and segmentation.}
    \label{fig:yoloe_concept}
\end{figure}

YOLOE-26 represents a significant evolution in the lineage by integrating the high-performance YOLO26 architecture with advanced open-vocabulary capabilities. By aligning visual features with rich linguistic embeddings, this capability enables the real-time detection and instance segmentation of arbitrary object classes, effectively removing the historical constraints of fixed-category training \cite{radford2021learning}. The framework provides flexible inference options to adapt to dynamic scenarios. As illustrated conceptually in Figure \ref{fig:yoloe_concept}, YOLOE-26 supports three distinct modes: utilizing text prompts to define targets (e.g., "find the red cup"), employing visual prompts via reference images for one-shot recognition, or operating in a prompt-free mode for zero-shot inference.

\begin{figure}[h!]
    \centering
    \includegraphics[width=\textwidth]{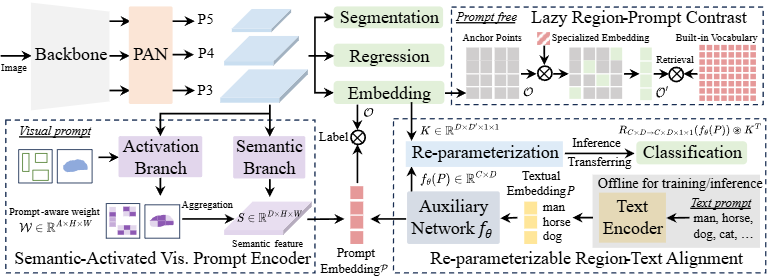} 
    \caption{Detailed architecture pipeline of YOLOE. The diagram illustrates how the RepRTA, SAVPE, and LRPC modules interface with the standard feature extraction backbone and NMS-free decoupled head.}
    \label{fig:yoloe_arch}
\end{figure}

To achieve these flexible multi-modal inputs without bottlenecking real-time edge performance, YOLOE-26 structurally modifies the standard YOLO backbone and PAN-FPN neck by introducing three novel modules, detailed in Figure \ref{fig:yoloe_arch}:

\begin{itemize}
    \item \textbf{Re-parameterizable Region-Text Alignment (RepRTA):} Refines text embeddings (e.g., from CLIP) via a small auxiliary network to support text-prompted detection.
    \item \textbf{Semantic-Activated Visual Prompt Encoder (SAVPE):} Encodes semantic and activation features from a reference image, conditioning the model for one-shot visual prompting.
    \item \textbf{Lazy Region-Prompt Contrast (LRPC):} Enables prompt-free zero-shot inference by performing open-set recognition using internal embeddings trained on massive vocabularies (e.g., LVIS \cite{gupta2019lvis}, Objects365 \cite{shao2019objects365}).
\end{itemize}

\vspace{-2mm}
The primary architectural motivation behind this specific pipeline is \textit{zero-overhead inference}. Post-training, the parameters of the RepRTA and SAVPE modules can be re-parameterized and folded directly into a standard YOLO head. As a result, when utilized as a regular closed-set detector, YOLOE-26 preserves identical FLOPs and latency to standard YOLO26 models. 

From a technical perspective, YOLOE-26 continues to leverage the native \textbf{NMS-free, end-to-end design} of the core backbone. This design eliminates the need for heuristic post-processing steps like Non-Maximum Suppression (NMS), a paradigm shift popularized by transformer-based detectors \cite{carion2020end}. By building upon this streamlined architecture, the model delivers fast open-world inference with minimal latency. This combination of deterministic high-speed performance and semantic flexibility makes YOLOE-26 a powerful solution for edge applications deployed in environments where the objects of interest represent a broad and evolving vocabulary \cite{yolo26_ultralytics}.

\section{Official Performance Benchmarks and Analysis}
\label{sec:benchmarks}
\vspace{-2mm}
To quantify the impact of the architectural innovations discussed in previous sections, this study reviews the official performance metrics published by the Ultralytics development team \cite{yolo26_ultralytics}. The following benchmarks evaluate YOLO26 on standard datasets, validating the efficacy of the NMS-free, end-to-end architecture.

\subsection{Object Detection Performance}
As detailed in the official YOLO26 documentation, the baseline evaluations were conducted on the Microsoft COCO \texttt{val2017} dataset, which includes 80 pretrained classes. The most significant indicator of YOLO26's architectural efficiency is its performance on standard object detection tasks. Table \ref{tab:det_benchmark} presents the official metrics for the YOLO26 family, ranging from the highly constrained Nano (n) variant to the Extra-Large (x) model.

\begin{table}[h!]
\centering
\caption{Official YOLO26 Object Detection Benchmarks on COCO. Hardware speeds represent CPU (ONNX) and NVIDIA T4 GPU (TensorRT10) environments.}
\label{tab:det_benchmark}
\resizebox{\textwidth}{!}{%
\begin{tabular}{lcccccccc}
\toprule
\textbf{Model} & \textbf{Size (px)} & \textbf{mAP$^{\text{val}}_{\text{50-95}}$} & \textbf{mAP$^{\text{val}}_{\text{e2e}}$} & \textbf{CPU ONNX (ms)} & \textbf{T4 TensorRT (ms)} & \textbf{Params (M)} & \textbf{FLOPs (B)} \\ \midrule
YOLO26n & 640 & 40.9 & 40.1 & 38.9 $\pm$ 0.7 & 1.7 $\pm$ 0.0 & 2.4 & 5.4 \\
YOLO26s & 640 & 48.6 & 47.8 & 87.2 $\pm$ 0.9 & 2.5 $\pm$ 0.0 & 9.5 & 20.7 \\
YOLO26m & 640 & 53.1 & 52.5 & 220.0 $\pm$ 1.4 & 4.7 $\pm$ 0.1 & 20.4 & 68.2 \\
YOLO26l & 640 & 55.0 & 54.4 & 286.2 $\pm$ 2.0 & 6.2 $\pm$ 0.2 & 24.8 & 86.4 \\
YOLO26x & 640 & 57.5 & 56.9 & 525.8 $\pm$ 4.0 & 11.8 $\pm$ 0.2 & 55.7 & 193.9 \\ \bottomrule
\end{tabular}%
}
\end{table}
\vspace{-2mm}
\noindent\footnotesize{\textit{*Note: $\text{mAP}^{\text{val}}$ metrics represent single-model, single-scale evaluation on the COCO \texttt{val2017} dataset. Parameter counts and FLOPs denote the fused inference architecture, which merges Conv/BatchNorm layers and removes the auxiliary one-to-many detection head used during training.}}
\normalsize

As reported in the official metrics, YOLO26 establishes a strict Pareto dominance across all model scales. Notably, the $\text{mAP}^{\text{val}}_{\text{e2e}}$ column confirms that the native end-to-end architecture retains nearly all the precision of the standard evaluation baseline, while the latency columns demonstrate the extreme computational efficiency of the framework. The Nano variant (YOLO26n), for instance, achieves over 40 mAP at a negligible 1.7 ms latency on a T4 GPU, rendering it highly competitive for real-world edge deployment.

\subsection{Instance Segmentation Performance}
Beyond standard bounding boxes, the official release includes benchmarks for pixel-level instance segmentation. Because YOLO26 utilizes a unified, DFL-free backbone across all its tasks, the computational penalty typically associated with adding a mask-prediction branch is heavily mitigated. Table \ref{tab:seg_benchmark} presents the performance of the \texttt{-seg} models on the COCO dataset.

\begin{table}[h!]
\centering
\caption{Official YOLO26 Instance Segmentation Benchmarks on COCO.}
\label{tab:seg_benchmark}
\resizebox{\textwidth}{!}{%
\begin{tabular}{lcccccccc}
\toprule
\textbf{Model} & \textbf{Size (px)} & \textbf{mAP$^{\text{box}}_{\text{e2e}}$} & \textbf{mAP$^{\text{mask}}_{\text{e2e}}$} & \textbf{CPU ONNX (ms)} & \textbf{T4 TensorRT (ms)} & \textbf{Params (M)} & \textbf{FLOPs (B)} \\ \midrule
YOLO26n-seg & 640 & 39.6 & 33.9 & 53.3 $\pm$ 0.5 & 2.1 $\pm$ 0.0 & 2.7 & 9.1 \\
YOLO26s-seg & 640 & 47.3 & 40.0 & 118.4 $\pm$ 0.9 & 3.3 $\pm$ 0.0 & 10.4 & 34.2 \\
YOLO26m-seg & 640 & 52.5 & 44.1 & 328.2 $\pm$ 2.4 & 6.7 $\pm$ 0.1 & 23.6 & 121.5 \\
YOLO26l-seg & 640 & 54.4 & 45.5 & 387.0 $\pm$ 3.7 & 8.0 $\pm$ 0.1 & 28.0 & 139.8 \\
YOLO26x-seg & 640 & 56.5 & 47.0 & 787.0 $\pm$ 6.8 & 16.4 $\pm$ 0.1 & 62.8 & 313.5 \\ \bottomrule
\end{tabular}%
}
\end{table}

The data illustrates that high-fidelity contour extraction is exportable to real-time edge environments. For example, the YOLO26n-seg model requires only 2.7M parameters—a marginal increase over the base detection model—yet achieves nearly 34.0 mask mAP with a T4 inference latency of just 2.1 ms. This validates the efficacy of the ProgLoss "contour polishing" dynamic discussed in Section \ref{sec:tasks}.

\subsection{Image Classification Performance}
For image classification, the YOLO26 architecture prioritizes holistic visual reasoning via Global Average Pooling (GAP). The official benchmarks assess this capability on the ImageNet dataset, evaluating the model across 1000 pretrained classes at a standard resolution of 224x224 pixels.

\begin{table}[h!]
\centering
\caption{Official YOLO26 Image Classification Benchmarks on ImageNet.}
\label{tab:cls_benchmark}
\resizebox{\textwidth}{!}{%
\begin{tabular}{lcccccccc}
\toprule
\textbf{Model} & \textbf{Size (px)} & \textbf{Acc Top-1} & \textbf{Acc Top-5} & \textbf{CPU ONNX (ms)} & \textbf{T4 TensorRT (ms)} & \textbf{Params (M)} & \textbf{FLOPs (B)} \\ \midrule
YOLO26n-cls & 224 & 71.4 & 90.1 & 5.0 $\pm$ 0.3 & 1.1 $\pm$ 0.0 & 2.8 & 0.5 \\
YOLO26s-cls & 224 & 76.0 & 92.9 & 7.9 $\pm$ 0.2 & 1.3 $\pm$ 0.0 & 6.7 & 1.6 \\
YOLO26m-cls & 224 & 78.1 & 94.2 & 17.2 $\pm$ 0.4 & 2.0 $\pm$ 0.0 & 11.6 & 4.9 \\
YOLO26l-cls & 224 & 79.0 & 94.6 & 23.2 $\pm$ 0.3 & 2.8 $\pm$ 0.0 & 14.1 & 6.2 \\
YOLO26x-cls & 224 & 79.9 & 95.0 & 41.4 $\pm$ 0.9 & 3.8 $\pm$ 0.0 & 29.6 & 13.6 \\ \bottomrule
\end{tabular}%
}
\end{table}

The YOLO26x-cls model achieves nearly 80.0\% Top-1 accuracy while maintaining a sub-4-millisecond inference speed on a T4 GPU, making it a highly optimal feature extractor for upstream perception pipelines.

\subsection{Pose Estimation Performance}
Designated by the \texttt{-pose} suffix (e.g., \texttt{yolo26n-pose.pt}), these variants are explicitly trained on the COCO keypoints dataset, rendering them highly adaptable for a wide variety of downstream pose estimation tasks. The official performance metrics for these models are detailed in Table \ref{tab:pose_benchmark}, demonstrating the effectiveness of the NMS-free architecture for precise spatial tracking.

\begin{table}[h!]
\centering
\caption{Official YOLO26 Pose Estimation Benchmarks on COCO.}
\label{tab:pose_benchmark}
\resizebox{\textwidth}{!}{%
\begin{tabular}{lcccccccc}
\toprule
\textbf{Model} & \textbf{Size (px)} & \textbf{mAP$^{\text{pose}}_{\text{50-95(e2e)}}$} & \textbf{mAP$^{\text{pose}}_{\text{50(e2e)}}$} & \textbf{CPU ONNX (ms)} & \textbf{T4 TensorRT (ms)} & \textbf{Params (M)} & \textbf{FLOPs (B)} \\ \midrule
YOLO26n-pose & 640 & 57.2 & 83.3 & 40.3 $\pm$ 0.5 & 1.8 $\pm$ 0.0 & 2.9 & 7.5 \\
YOLO26s-pose & 640 & 63.0 & 86.6 & 85.3 $\pm$ 0.9 & 2.7 $\pm$ 0.0 & 10.4 & 23.9 \\
YOLO26m-pose & 640 & 68.8 & 89.6 & 218.0 $\pm$ 1.5 & 5.0 $\pm$ 0.1 & 21.5 & 73.1 \\
YOLO26l-pose & 640 & 70.4 & 90.5 & 275.4 $\pm$ 2.4 & 6.5 $\pm$ 0.1 & 25.9 & 91.3 \\
YOLO26x-pose & 640 & 71.6 & 91.6 & 565.4 $\pm$ 3.0 & 12.2 $\pm$ 0.2 & 57.6 & 201.7 \\ \bottomrule
\end{tabular}%
}
\end{table}

These findings confirm that the removal of Distribution Focal Loss (DFL) does not degrade spatial keypoint tracking. The YOLO26n-pose model yields an impressive 57.2 $\text{mAP}^{\text{pose}}_{\text{50-95(e2e)}}$ at 1.8 ms (T4 TensorRT), verifying its suitability for real-time biomechanical analysis on edge devices.

\subsection{Oriented Bounding Boxes (OBB) Object Detection Performance}
For tasks involving skewed or densely packed targets (e.g., aerial or satellite imagery), the OBB models are evaluated on the DOTAv1 dataset across 15 pretrained classes. Due to the requirement for higher spatial fidelity in aerial contexts, these models operate at a larger inference resolution of 1024x1024 pixels.

\begin{table}[h!]
\centering
\caption{Official YOLO26 Oriented Object Detection Benchmarks on DOTAv1.}
\label{tab:obb_benchmark}
\resizebox{\textwidth}{!}{%
\begin{tabular}{lcccccccc}
\toprule
\textbf{Model} & \textbf{Size (px)} & \textbf{mAP$^{\text{test}}_{\text{50-95(e2e)}}$} & \textbf{mAP$^{\text{test}}_{\text{50(e2e)}}$} & \textbf{CPU ONNX (ms)} & \textbf{T4 TensorRT (ms)} & \textbf{Params (M)} & \textbf{FLOPs (B)} \\ \midrule
YOLO26n-obb & 1024 & 52.4 & 78.9 & 97.7 $\pm$ 0.9 & 2.8 $\pm$ 0.0 & 2.5 & 14.0 \\
YOLO26s-obb & 1024 & 54.8 & 80.9 & 218.0 $\pm$ 1.4 & 4.9 $\pm$ 0.1 & 9.8 & 55.1 \\
YOLO26m-obb & 1024 & 55.3 & 81.0 & 579.2 $\pm$ 3.8 & 10.2 $\pm$ 0.3 & 21.2 & 183.3 \\
YOLO26l-obb & 1024 & 56.2 & 81.6 & 735.6 $\pm$ 3.1 & 13.0 $\pm$ 0.2 & 25.6 & 230.0 \\
YOLO26x-obb & 1024 & 56.7 & 81.7 & 1485.7 $\pm$ 11.5 & 30.5 $\pm$ 0.9 & 57.6 & 516.5 \\ \bottomrule
\end{tabular}%
}
\end{table}

Despite the computationally intensive 1024x1024 input resolution, the NMS-free architecture ensures latency remains strictly bounded. The YOLO26s-obb variant processes these large matrices in under 5.0 ms on a T4 GPU, effectively resolving the boundary discontinuity errors typically associated with high-resolution aerial detection.

\subsection{Open-Vocabulary Instance Segmentation (YOLOE-26)}
To validate the multi-modal capabilities discussed in Section \ref{subsec:openvoc}, the official benchmarks assess the YOLOE-26 series on open-vocabulary detection and segmentation. These models are evaluated using a combination of the Objects365v1 \cite{shao2019objects365}, GQA \cite{hudson2019gqa}, and Flickr30k \cite{plummer2015flickr30k} datasets.
Table \ref{tab:yoloe_benchmark} presents the performance metrics utilizing both Text and Visual prompts. The metrics are denoted in a (Text / Visual) format to illustrate the model's flexibility across different prompting modalities.

\begin{table}[h!]
\centering
\caption{Official YOLOE-26 Open-Vocabulary Instance Segmentation Benchmarks. Performance values are reported as (Text Prompt / Visual Prompt).}
\label{tab:yoloe_benchmark}
\resizebox{\textwidth}{!}{%
\begin{tabular}{lcccccccc}
\toprule
\textbf{Model} & \textbf{Size (px)} & \textbf{mAP$^{\text{minival}}_{\text{50-95(e2e)}}$} & \textbf{mAP$^{\text{minival}}_{\text{50-95}}$} & \textbf{mAP$_r$} & \textbf{mAP$_c$} & \textbf{mAP$_f$} & \textbf{Params (M)} & \textbf{FLOPs (B)} \\ \midrule
YOLOE-26n-seg & 640 & 23.7 / 20.9 & 24.7 / 21.9 & 20.5 / 17.6 & 24.1 / 22.3 & 26.1 / 22.4 & 4.8 & 6.0 \\
YOLOE-26s-seg & 640 & 29.9 / 27.1 & 30.8 / 28.6 & 23.9 / 25.1 & 29.6 / 27.8 & 33.0 / 29.9 & 13.1 & 21.7 \\
YOLOE-26m-seg & 640 & 35.4 / 31.3 & 35.4 / 33.9 & 31.1 / 33.4 & 34.7 / 34.0 & 36.9 / 33.8 & 27.9 & 70.1 \\
YOLOE-26l-seg & 640 & 36.8 / 33.7 & 37.8 / 36.3 & 35.1 / 37.6 & 37.6 / 36.2 & 38.5 / 36.1 & 32.3 & 88.3 \\
YOLOE-26x-seg & 640 & 39.5 / 36.2 & 40.6 / 38.5 & 37.4 / 35.3 & 40.9 / 38.8 & 41.0 / 38.8 & 69.9 & 196.7 \\ \bottomrule
\end{tabular}%
}
\end{table}

The data reveals the architectural overhead required to align visual features with rich linguistic embeddings. While the parameter count naturally increases compared to the standard segmentation models (e.g., YOLOE-26n-seg requires 4.8M parameters versus 2.7M for YOLO26n-seg), the end-to-end framework efficiently manages this multi-modal complexity. The models maintain strong recall across rare ($\text{mAP}_r$), common ($\text{mAP}_c$), and frequent ($\text{mAP}_f$) classes, confirming that the NMS-free design is highly scalable to open-world, dynamic environments where fixed-category constraints are removed.

Furthermore, the YOLOE-26 framework offers a "Prompt-Free" (zero-shot) mode, designed for autonomous environments where external text or visual prompts are unavailable. Table \ref{tab:yoloe_pf_benchmark} details the performance of the specialized \texttt{-pf} (prompt-free) variants.

\begin{table}[h!]
\centering
\caption{YOLOE-26 Prompt-Free (Zero-Shot) Benchmarks on Objects365v1, GQA, and Flickr30k.}
\label{tab:yoloe_pf_benchmark}
\resizebox{0.9\textwidth}{!}{%
\begin{tabular}{lccccc}
\toprule
\textbf{Model} & \textbf{Size (px)} & \textbf{mAP$^{\text{minival}}_{\text{50-95(e2e)}}$} & \textbf{mAP$^{\text{minival}}_{\text{50(e2e)}}$} & \textbf{Params (M)} & \textbf{FLOPs (B)} \\ \midrule
YOLOE-26n-seg-pf & 640 & 16.6 & 22.7 & 6.5 & 15.8 \\
YOLOE-26s-seg-pf & 640 & 21.4 & 28.6 & 16.2 & 35.5 \\
YOLOE-26m-seg-pf & 640 & 25.7 & 33.6 & 36.2 & 122.1 \\
YOLOE-26l-seg-pf & 640 & 27.2 & 35.4 & 40.6 & 140.4 \\
YOLOE-26x-seg-pf & 640 & 29.9 & 38.7 & 86.3 & 314.4 \\ \bottomrule
\end{tabular}%
}
\end{table}

While operating without explicit guidance naturally results in lower overall mAP compared to prompt-assisted modes, the prompt-free models retain substantial zero-shot detection capabilities. The corresponding increase in parameter and computational load—for instance, the Nano prompt-free variant requires 6.5M parameters and 15.8B FLOPs compared to 4.8M and 6.0B for the prompted version—reflects the heavier internal encoding required to independently reason about open-world object classes without an external semantic anchor.

\subsection{Comprehensive State-of-the-Art Analysis}
To establish the efficacy of the YOLO26 architecture, we conduct an exhaustive comparison against the current State-of-the-Art (SOTA) object detection models. The benchmark data, sourced from the Roboflow Computer Vision Leaderboard\footnote{Leaderboard: \url{https://leaderboard.roboflow.com/}}, evaluates models on the COCO \texttt{val2017} dataset. This comparison spans the entire spectrum of model scales, from highly constrained Nano variants to high-capacity Extra-Large models, incorporating both CNN-based YOLO lineages (v8 through v13) and recent Transformer-based architectures (RT-DETR, DEIM, RF-DETR).

\begin{center}
\small
\setlength{\LTcapwidth}{\textwidth}
\begin{longtable}{lccccccc}
\caption{Comprehensive SOTA Object Detection Benchmarks on COCO \texttt{val2017}. Models are grouped by scale and sorted by mAP$^{\text{val}}_{\text{50-95}}$. All performance metrics (mAP and F1 scores) are reported as percentages (\%).} \label{tab:full_sota} \\
\toprule
\textbf{Model} & \textbf{Params (M)} & \textbf{mAP$^{\text{val}}_{\text{50-95}}$} & \textbf{mAP$^{\text{val}}_{\text{50}}$} & \textbf{mAP$^{\text{val}}_{\text{75}}$} & \textbf{F1$_{\text{50}}$} & \textbf{F1$_{\text{75}}$} & \textbf{License} \\ 
\midrule
\endfirsthead

\multicolumn{8}{c}%
{{\bfseries \tablename\ \thetable{} -- continued from previous page}} \\
\toprule
\textbf{Model} & \textbf{Params (M)} & \textbf{mAP$^{\text{val}}_{\text{50-95}}$} & \textbf{mAP$^{\text{val}}_{\text{50}}$} & \textbf{mAP$^{\text{val}}_{\text{75}}$} & \textbf{F1$_{\text{50}}$} & \textbf{F1$_{\text{75}}$} & \textbf{License} \\ 
\midrule
\endhead

\midrule \multicolumn{8}{r}{{Continued on next page}} \\ \midrule
\endfoot

\bottomrule
\endlastfoot

\multicolumn{8}{c}{\textbf{Large \& Extra-Large Models}} \\ \midrule
RF-DETR-XXL & 126.9 & 59.9 & 78.2 & 65.4 & 15.3 & 12.9 & PML-1.0 \\
RF-DETR-XL & 126.4 & 58.5 & 77.1 & 63.7 & 15.0 & 12.4 & PML-1.0 \\
DEIM-D-FINE-X & 61.7 & 56.5 & 74.0 & 61.6 & 5.7 & 4.8 & Apache-2.0 \\
\textbf{YOLO26x} & \textbf{55.7} & \textbf{56.3} & \textbf{73.4} & \textbf{61.7} & \textbf{14.4} & \textbf{12.5} & \textbf{AGPL-3.0} \\
RF-DETR-L & 33.9 & 56.3 & 74.8 & 61.1 & 15.2 & 12.6 & Apache-2.0 \\
DEIM-RT-DETRv2-X & 74.9 & 55.5 & 73.5 & 60.3 & 5.7 & 4.7 & Apache-2.0 \\
DEIM-D-FINE-L & 30.8 & 54.7 & 72.4 & 59.4 & 5.6 & 4.7 & Apache-2.0 \\
RT-DETRv2-X & 92.5 & 54.3 & 72.8 & 58.8 & 5.6 & 4.6 & Apache-2.0 \\
RT-DETR-R101 & 92.5 & 54.3 & 72.8 & 58.8 & 5.6 & 4.6 & Apache-2.0 \\
DEIM-RT-DETRv2-L & 42.1 & 54.3 & 72.2 & 58.8 & 5.7 & 4.7 & Apache-2.0 \\
YOLOv12x & 59.1 & 54.0 & 70.3 & 59.0 & 26.2 & 22.5 & AGPL-3.0 \\
YOLOv13x & 64.0 & 53.7 & 70.8 & 58.7 & 13.8 & 11.6 & AGPL-3.0 \\
YOLOv10x & 31.7 & 53.6 & 70.3 & 58.4 & 22.6 & 19.4 & AGPL-3.0 \\
YOLO11x & 56.9 & 53.6 & 70.2 & 58.4 & 13.9 & 11.8 & AGPL-3.0 \\
\textbf{YOLO26l} & \textbf{24.8} & \textbf{53.6} & \textbf{70.4} & \textbf{58.6} & \textbf{14.2} & \textbf{12.2} & \textbf{AGPL-3.0} \\
RT-DETRv2-L & 50.0 & 53.4 & 71.6 & 57.5 & 5.6 & 4.6 & Apache-2.0 \\
RT-DETR-R50 & 50.0 & 53.1 & 71.2 & 57.7 & 5.5 & 4.5 & Apache-2.0 \\
YOLOv8x & 68.2 & 52.9 & 69.4 & 57.7 & 23.6 & 20.0 & AGPL-3.0 \\
YOLOv12l & 26.4 & 52.6 & 69.1 & 57.3 & 24.4 & 20.7 & AGPL-3.0 \\
RTMDet-x & 94.9 & 52.5 & 70.1 & 57.7 & 5.4 & 4.5 & GPL-3.0 \\
YOLOv10l & 25.8 & 52.3 & 69.1 & 57.1 & 22.6 & 19.2 & AGPL-3.0 \\
YOLOv13l & 27.6 & 52.3 & 69.7 & 56.9 & 13.8 & 11.5 & AGPL-3.0 \\
YOLO11l & 25.3 & 52.2 & 68.5 & 56.9 & 13.7 & 11.5 & AGPL-3.0 \\
YOLOv8l & 43.7 & 51.8 & 68.3 & 56.5 & 22.8 & 19.3 & AGPL-3.0 \\
RTMDet-l & 52.3 & 51.2 & 68.9 & 55.8 & 5.5 & 4.5 & GPL-3.0 \\ \midrule

\multicolumn{8}{c}{\textbf{Medium Models}} \\ \midrule
RF-DETR-M & 33.7 & 54.8 & 73.6 & 59.3 & 5.7 & 4.6 & Apache-2.0 \\
DEIM-RT-DETRv2-M* & 33.0 & 53.2 & 71.2 & 57.8 & 5.7 & 4.6 & Apache-2.0 \\
DEIM-D-FINE-M & 19.2 & 52.7 & 70.0 & 57.3 & 5.6 & 4.6 & Apache-2.0 \\
\textbf{YOLO26m} & \textbf{20.4} & \textbf{52.0} & \textbf{69.0} & \textbf{56.8} & \textbf{14.1} & \textbf{12.0} & \textbf{AGPL-3.0} \\
RT-DETRv2-M* & 38.4 & 51.9 & 69.9 & 56.5 & 5.6 & 4.6 & Apache-2.0 \\
YOLOv10b & 20.5 & 51.8 & 68.6 & 56.6 & 22.0 & 18.6 & AGPL-3.0 \\
YOLOv12m & 20.2 & 51.4 & 68.0 & 55.8 & 23.5 & 19.8 & AGPL-3.0 \\
DEIM-RT-DETRv2-M & 31.2 & 50.9 & 68.6 & 55.2 & 5.6 & 4.6 & Apache-2.0 \\
YOLO11m & 20.1 & 50.5 & 67.1 & 55.0 & 13.5 & 11.3 & AGPL-3.0 \\
YOLOv10m & 16.5 & 50.3 & 67.2 & 54.9 & 21.2 & 17.8 & AGPL-3.0 \\
RT-DETRv2-M & 33.2 & 49.9 & 67.5 & 54.1 & 5.5 & 4.5 & Apache-2.0 \\
YOLOv8m & 25.9 & 49.2 & 65.7 & 53.6 & 20.2 & 16.8 & AGPL-3.0 \\
RTMDet-m & 24.7 & 49.0 & 66.7 & 53.6 & 5.5 & 4.4 & GPL-3.0 \\ \midrule

\multicolumn{8}{c}{\textbf{Small Models}} \\ \midrule
RF-DETR-S & 32.1 & 53.0 & 72.1 & 57.3 & 5.6 & 4.4 & Apache-2.0 \\
DEIM-RT-DETRv2-S & 20.0 & 49.1 & 66.1 & 53.3 & 5.7 & 4.6 & Apache-2.0 \\
DEIM-D-FINE-S & 10.2 & 49.0 & 65.9 & 53.1 & 5.5 & 4.5 & Apache-2.0 \\
RT-DETR-R34 & 33.2 & 48.9 & 66.8 & 52.7 & 5.4 & 4.4 & Apache-2.0 \\
RT-DETRv2-S & 22.0 & 48.1 & 65.1 & 52.1 & 5.5 & 4.5 & Apache-2.0 \\
\textbf{YOLO26s} & \textbf{9.5} & \textbf{47.2} & \textbf{63.5} & \textbf{51.5} & \textbf{13.5} & \textbf{11.1} & \textbf{AGPL-3.0} \\
YOLOv13s & 9.0 & 46.8 & 63.5 & 50.6 & 13.0 & 10.4 & AGPL-3.0 \\
YOLOv12s & 9.3 & 46.7 & 63.1 & 50.6 & 20.7 & 16.9 & AGPL-3.0 \\
RT-DETR-R18 & 22.0 & 46.4 & 63.7 & 50.3 & 5.4 & 4.3 & Apache-2.0 \\
YOLOv10s & 8.1 & 45.7 & 62.3 & 49.8 & 18.6 & 15.2 & AGPL-3.0 \\
YOLO11s & 9.4 & 45.5 & 62.0 & 49.3 & 12.8 & 10.4 & AGPL-3.0 \\
RTMDet-s & 8.9 & 44.4 & 61.5 & 48.1 & 5.0 & 3.9 & GPL-3.0 \\
YOLOv8s & 11.2 & 44.1 & 60.2 & 47.7 & 17.6 & 14.1 & AGPL-3.0 \\ \midrule

\multicolumn{8}{c}{\textbf{Nano \& Tiny Models}} \\ \midrule
DEIM-D-FINE-N & 10.2 & 49.0 & 65.9 & 53.1 & 5.5 & 4.5 & Apache-2.0 \\
RF-DETR-N & 30.5 & 48.4 & 67.5 & 51.8 & 5.2 & 3.9 & Apache-2.0 \\
RTMDet-t & 4.9 & 41.0 & 57.4 & 44.3 & 5.1 & 3.9 & GPL-3.0 \\
YOLOv13n & 2.5 & 40.4 & 56.2 & 43.9 & 12.2 & 9.4 & AGPL-3.0 \\
\textbf{YOLO26n} & \textbf{2.4} & \textbf{39.9} & \textbf{55.2} & \textbf{43.4} & \textbf{12.2} & \textbf{9.5} & \textbf{AGPL-3.0} \\
YOLOv12n & 2.6 & 39.7 & 55.0 & 42.9 & 16.7 & 13.1 & AGPL-3.0 \\
YOLO11n & 2.6 & 38.6 & 53.9 & 42.0 & 12.1 & 9.3 & AGPL-3.0 \\
YOLOv10n & 2.8 & 38.0 & 52.9 & 41.3 & 15.4 & 12.1 & AGPL-3.0 \\
YOLOv8n & 3.2 & 36.5 & 51.4 & 39.8 & 14.8 & 11.5 & AGPL-3.0 \\

\end{longtable}
\end{center}

As evidenced by the exhaustive data in Table \ref{tab:full_sota}, the YOLO26 architecture establishes a highly dominant Pareto frontier across all model scales, successfully bridging the gap between lightweight CNN efficiency and heavy Transformer accuracy. At the high-capacity end, the \textbf{YOLO26x} variant achieves an exceptional 56.3\% $\text{mAP}^{\text{val}}_{\text{50-95}}$ with only 55.7M parameters. This firmly eclipses contemporary heavyweight models such as YOLO11x (53.6\% mAP at 56.9M parameters) and fiercely rivals advanced Transformer architectures like DEIM-D-FINE-X (56.5\% mAP at 61.7M parameters), delivering comparable spatial reasoning without the massive computational overhead typical of self-attention mechanisms.

Furthermore, this architectural efficiency seamlessly cascades down to the most constrained edge environments. The \textbf{YOLO26n} model (2.4M parameters) secures 39.9\% mAP, outperforming equivalently scaled variants like YOLOv12n and YOLO11n. Across the entire YOLO26 family, the robust $\text{F1}_{\text{50}}$ and $\text{F1}_{\text{75}}$ scores demonstrate a superior precision-recall balance. This validates the core premise of the end-to-end, NMS-free design: by eliminating heuristic post-processing, the model fundamentally reduces false positives and boundary discontinuities, resulting in a sharper, more reliable perception pipeline.

\section{Implications for Edge AI: Bridging the "Export Gap"}
\label{sec:impact}
\vspace{-2mm}
A pervasive challenge in the modern era of object detection is the "Export Gap"—the discrepancy between the theoretical performance observed during GPU training and the actual latency realized on deployed edge hardware \cite{lyu2022rtmdet}. This section analyzes how YOLO26 addresses this critical bottleneck through its architectural constraints.

\subsection{The Latency Bottleneck in Traditional Models}
\vspace{-2mm}
Prior State-of-the-Art (SOTA) models, including YOLOv8 through YOLOv13, relied heavily on Distribution Focal Loss (DFL) to maximize mAP \cite{jocher2023yolov8, wang2024yolov10, lei2025yolov13}. While mathematically precise, DFL necessitates complex Softmax operations over discretized bins to calculate final coordinates \cite{li2020generalized}. On server-grade GPUs, these operations are negligible. However, on integer-arithmetic hardware (such as NPUs in mobile devices or DSPs in drones), Softmax layers are difficult to quantize and often become the primary latency bottleneck \cite{gholami2021survey}. Consequently, a model that appears efficient in a research paper often suffers severe throughput degradation when exported to real-world embedded systems.

\subsection{Deterministic Inference via Direct Regression}
\vspace{-2mm}
YOLO26 resolves this trade-off by reverting to a Direct Regression strategy, explicitly removing the computational burden of DFL \cite{jocher2026yolo26}. By decoupling representation learning from complex post-processing, the architecture ensures that the inference graph consists solely of standard convolutional and linear operations. This shift guarantees deterministic latency—the inference time remains constant regardless of scene complexity or object density \cite{jocher2026yolo26, lyu2022rtmdet}. This predictability is paramount for safety-critical edge applications, such as autonomous driving and robotic navigation, where timing violations can lead to catastrophic failures \cite{jocher2026yolo26}.

\section{Future Directions}
\label{sec:future_work}
\vspace{-2mm}
While YOLO26 establishes a new benchmark for real-time detection, several avenues for exploration remain to fully bridge the gap between edge efficiency and cognitive intelligence.

\textbf{Inherent Explainability and Trustworthiness:}
Currently, the "black box" nature of deep detectors is addressed via post-hoc methods like Grad-CAM \cite{selvaraju2017grad} or SHAP \cite{lundberg2017unified}, which approximate the model's decision-making process after inference. A critical future direction is the development of Inherent Explainability \cite{chakrabartyxai}, where the detection head outputs not only the bounding box and class but also a justification map or textual rationale (e.g., "Classified as \textit{Tumor} due to irregular border texture"). Embedding interpretability directly into the end-to-end pipeline will be transformative for safety-critical domains such as medical diagnostics and autonomous defense, ensuring that high-speed decisions are also transparent and verifiable.

\textbf{Unified Spatiotemporal Perception:}
The NMS-free, deterministic nature of YOLO26 makes it uniquely suited for video analysis. Traditional detectors often suffer from "flicker" in video streams because NMS arbitrarily selects different boxes across frames. Future iterations could extend the YOLO26 backbone to handle Spatiotemporal Object Detection natively. By treating time as a third spatial dimension, the model could perform tracking and action recognition (e.g., "person running") within the same single-pass forward pass, eliminating the need for separate tracking algorithms like DeepSORT \cite{wojke2017simple}.

\textbf{Test-Time Adaptation on the Edge:}
Finally, the static nature of trained models remains a limitation in dynamic environments. Future work should explore Test-Time Adaptation (TTA) \cite{sun2020test}, allowing the model to update its batch normalization statistics or lightweight adapter layers directly on the edge device. This would enable a drone or medical device to "acclimatize" to new lighting conditions or sensor noise profiles in real-time, maintaining peak accuracy without requiring a full retraining cycle on a server.

\section{Conclusion}
\label{sec:conclusion}
\vspace{-2mm}
This study presents a comprehensive analysis of the YOLO26 architecture, which advances the real-time object detection paradigm by eliminating Non-Maximum Suppression (NMS) in favor of a native end-to-end learning strategy. Supported by core innovations such as the MuSGD optimizer, Small-Target-Aware Label Assignment (STAL), and ProgLoss scheduling, this transition successfully resolves historical latency bottlenecks. Furthermore, the adoption of a Direct Regression head effectively closes the "Export Gap," ensuring deterministic latency for resource-constrained edge devices. As demonstrated by extensive benchmarking against prior YOLO lineages and contemporary Transformer architectures, YOLO26 establishes a dominant new speed-accuracy Pareto front. Additionally, the analysis of the YOLOE-26 open-vocabulary module highlights the framework's unified capacity for zero-overhead, promptable multi-task detection. Ultimately, by decoupling representation learning from heuristic post-processing, YOLO26 signals a fundamental shift toward fully learnable, hardware-aware pipelines, providing a highly reliable blueprint for the next generation of safety-critical Edge AI applications.

\section*{Acknowledgement(s)}
\vspace{-2mm}
The author explicitly acknowledges the use of Artificial Intelligence tools solely for the purpose of language refinement and grammatical polishing; all scientific concepts, data, and technical innovations presented herein are the original work of the author. All architectural interpretations and mathematical formulations are author-derived abstractions intended for conceptual clarity and do not represent official Ultralytics specifications. Official documentation is available at: \url{https://docs.ultralytics.com/models/yolo26/}.

\bibliographystyle{IEEEtran}
\bibliography{references}  

\end{document}